\documentclass[preprint,12pt]{elsarticle}
\biboptions{sort&compress}

\usepackage{xcolor}
\usepackage{graphicx}
\usepackage{upgreek}
\usepackage{epstopdf}
\usepackage{bm}

\usepackage{amsmath,amssymb}
\usepackage{booktabs}
\usepackage{tabularx}
\usepackage{longtable}
\usepackage{multirow}
\usepackage[varqu,scaled=0.95]{zi4}
\usepackage{listings}
\definecolor{codebg}{HTML}{F5F6F8}
\definecolor{coderule}{HTML}{B8C2CC}
\definecolor{codekw}{HTML}{2F78B7}
\definecolor{codestr}{HTML}{8A5F17}
\definecolor{codecom}{HTML}{6E6E6E}
\usepackage[hidelinks]{hyperref}
\usepackage[noabbrev]{cleveref}

\newif\ifshowrevisions
\showrevisionsfalse
\definecolor{revcolor}{HTML}{B23A48}

\newcommand{\revnote}[1]{\ifshowrevisions{\color{revcolor}\small[#1]\par}\fi}
\definecolor{revbcolor}{HTML}{2F78B7}

\newcommand{\revnoteb}[1]{\ifshowrevisions{\color{revbcolor}\small[#1]\par}\fi}
\crefname{figure}{Fig.}{Figs.}
\Crefname{figure}{Figure}{Figures}
\crefname{equation}{Eq.}{Eqs.}
\Crefname{equation}{Equation}{Equations}
\crefname{table}{Table}{Tables}
\Crefname{table}{Table}{Tables}
\crefname{section}{Section}{Sections}
\Crefname{section}{Section}{Sections}
\crefname{subsection}{Section}{Sections}
\Crefname{subsection}{Section}{Sections}
\crefformat{appendix}{#2#1#3}
\Crefformat{appendix}{#2#1#3}
\crefname{subappendix}{Appendix}{Appendices}
\Crefname{subappendix}{Appendix}{Appendices}

\graphicspath{{figures/}}



\journal{Journal of Computational Physics}

\begin{document}

\begin{frontmatter}

\title{LegoSR: A Unified Framework for Self-Improving Agent-Guided Scientific Discovery\tnoteref{code}}
\tnotetext[code]{Code is available at \href{https://github.com/scientific-artificial-intelligence-lab/KD}{\nolinkurl{github.com/scientific-artificial-intelligence-lab/KD}}.}

\author[eit,polyu]{Hao~Mao}
\author[uw]{Xu~T.~Liu}
\author[eit]{Hao~Xu}
\author[indep]{Peng~Zhao}
\author[eit,sjtu]{Qian~Li}
\author[eit]{Dongxiao~Zhang}
\author[polyu]{Jinyue~Yan}
\author[eit]{Yuntian~Chen\corref{cor1}}
\ead{ychen@eitech.edu.cn}
\cortext[cor1]{Corresponding author.}

\address[eit]{Eastern Institute of Technology, Ningbo, China}
\address[polyu]{The Hong Kong Polytechnic University, Hong Kong SAR, China}
\address[uw]{University of Washington, Seattle, USA}
\address[indep]{Independent Researcher, China}
\address[sjtu]{Shanghai Jiao Tong University, Shanghai, China}
\begin{abstract}
Discovering equations from data provides governing laws, constitutive relations and closure models for scientific understanding and prediction.
Yet a method that succeeds on one system may fail on another or on noisy data, so discovery depends on expert trial and error.
Unsuccessful searches can still reveal useful parts of an equation, but these clues must be recognized and combined.
We present LegoSR, a unified framework for agent-guided equation discovery built on composable discovery and cumulative experience.
Seven complementary equation discovery methods and shared computational modules form reusable building blocks, connected through a common equation representation.
Researchers can launch discovery with one command and guide it with partially known equations.
LegoSR Agent selects methods, revises searches using feedback and combines partial findings into new hypotheses.
Its Experience Library consolidates graded outcomes into lessons that guide later tasks across equation families.
Evaluation covers 134 algebraic regression problems, 40 ordinary differential equation targets and 25 partial differential equation problems at several noise levels, together with three case studies.
On noisy PDEs, LegoSR Agent recovers more equations than seven expert-tuned methods combined and stays close to them on ODEs.
It also exceeds a generic agent with the same model, tools and budget by up to 28 percentage points, with 63--73\% shorter median wall-clock time on PDE problems.
\end{abstract}


\end{frontmatter}

\section{Introduction}
\label{sec:introduction}

Discovering equations from data provides governing laws, constitutive relations and closure models for scientific understanding and prediction.
As observational and experimental data accumulate, data-driven equation discovery has become an active area of scientific computing and artificial intelligence~\cite{lou2026discoveryreview}.
Methods for symbolic regression (SR) and ordinary and partial differential equation (ODE/PDE) discovery include sparse regression~\cite{brunton2016sindy,rudy2017pdefind}, evolutionary search~\cite{xu2020dlga,chen2022sgapde,cranmer2023pysr}, reinforcement learning~\cite{du2024discover,du2024rdiscover} and language models~\cite{du2024llm4ed,xu2025eqgpt}.
These methods offer complementary strengths through different mathematical representations, search strategies and sources of prior knowledge (\cref{sec:related_individual}).
Sparse regression, for example, efficiently identifies equations within a suitable candidate library, whereas methods that build expressions can explore structures beyond a predefined library.

Yet a method that succeeds on one system can fail on another or on noisy data, so equation discovery depends on expert trial and error.
Researchers choose and configure methods, read intermediate results and decide whether to continue a search, adjust its configuration or switch to another method.
Benchmarks support systematic comparison of methods~\cite{lacava2021srbench} but do not determine which method and configuration to use as a search unfolds.
Failed searches still reveal useful parts of an equation, such as terms on which several searches agree, but a researcher must recognize these parts and combine them into a new hypothesis.
This trial and error is also laborious: most methods are released as research code, so using several of them means converting data and equation formats, keeping run records and restarting interrupted runs.
Lessons about which choices succeed or fail under given data conditions also remain scattered across experiments.
Each new problem therefore starts again with trial and error.

This is work for an agent: choosing and configuring methods, reading intermediate results, combining partial findings and carrying lessons to the next problem.
An agent that takes it over reduces reliance on experts and speeds up discovery.
Building such an agent is the aim of this paper.
Large language model (LLM) agents, which reason step by step and call software tools, can take on such work~\cite{yao2023react}.
Such agents have carried out chemical experiments~\cite{boiko2023coscientist}, generated research hypotheses~\cite{gottweis2026coscientist} and solved mechanics problems with finite element code~\cite{ni2024mechagents}.
If the methods share common interfaces, an agent can run any of them and reuse their results, but it must still decide which method to run and how to configure it.
Generic agents lack knowledge of equation discovery methods and of when each applies, and they start every task without the lessons of earlier ones.
The central challenge is therefore to turn isolated equation searches into an autonomous discovery process that combines complementary methods, revises hypotheses from feedback and accumulates transferable experience.

Adaptive coordination across equation discovery methods requires two complementary capabilities: composable discovery and cumulative experience.
Composable discovery lets the agent combine methods within one task: it can switch between them and pass equations, derivatives and partial findings from one search to the next.
This requires a common task definition, an equation representation that also covers partially known equations, shared computational modules and common evaluation, while each method keeps its own search mechanism.
Cumulative experience lets the agent build on earlier searches, both within a task and across tasks.
This requires a record of every search and lessons that tie the outcomes of method and configuration choices to data conditions, not to specific equations, so that they transfer to new equations.
Composable discovery gives the agent the means to act on many methods; cumulative experience tells it which actions suit the data at hand.

We present LegoSR, a unified open-source framework that combines these capabilities for algebraic symbolic regression and differential equation discovery.
The name reflects its building-block design: like Lego bricks, discovery methods, computational modules and partially known equations can be combined and reused to assemble workflows for different discovery tasks (\cref{fig:architecture}).
Seven complementary equation discovery methods and shared computational modules form reusable building blocks, connected through a common Equation Intermediate Representation (Equation IR).
A discovery task specifies the data and can include prior knowledge as a Sketch, a partially known equation.
LegoSR evaluates candidates from all methods with the same data processing and derivative estimation.
These blocks provide composable discovery.
Researchers can assemble them directly or launch LegoSR Agent with one command and guide it with partially known equations.
LegoSR also records every run and can resume interrupted ones, turning separate runs into a continuous and traceable research process.

LegoSR Agent assembles these blocks into a workflow for each task and takes over the decisions that otherwise rely on expert trial and error.
The agent draws on the papers that introduced the methods and on previous experiments to choose among them and configure searches.
Over successive iterations, it evaluates candidate equations and search progress, then decides whether to continue, adjust configurations or switch methods.
When searches fail, it builds on their results: it can combine the terms they agree on into a Sketch, a new hypothesis for the next search.
Its memory provides cumulative experience at two time scales.
Within a task, each step and its rationale enter a discovery tree that guides the rest of the task.
Across tasks, the Experience Library collects lessons from each graded task, so LegoSR Agent refines its guidance with every problem it attempts.
Evaluation covers 134 algebraic SR problems, 40 ODE problems and 25 PDE problems at several noise levels, together with three case studies.
On noisy PDE benchmarks, LegoSR Agent recovers more equations than all seven expert-tuned methods combined, and on ODEs it stays close to them.
It also outperforms a generic agent with the same model, tools and budget by up to 28 percentage points, with 63--73\% shorter median wall-clock time on PDE problems.

\section{Related work}
\label{sec:related-work}

\subsection{Individual equation discovery methods}
\label{sec:related_individual}

Early work recovered mathematical relationships from experimental data by evolutionary search~\cite{schmidt2009distilling}.
Sparse regression methods such as SINDy~\cite{brunton2016sindy} and PDE-FIND~\cite{rudy2017pdefind}, implemented in PySINDy~\cite{kaptanoglu2022pysindy}, select terms from predefined candidate libraries.
They are efficient but depend on the library and on derivative quality; weak and integral formulations reduce their sensitivity to noisy derivatives~\cite{messenger2021wsindy,xu2021integral}.

More flexible representations and search strategies reduce these dependencies.
AI Feynman~\cite{udrescu2020aifeynman} exploits symmetry and separability to simplify symbolic regression.
DLGA-PDE~\cite{xu2020dlga} and SGA-PDE~\cite{chen2022sgapde} build candidate equations by evolutionary search, so they need no complete library.
Deep Symbolic Regression~\cite{petersen2021dsr} generates expressions by reinforcement learning, and DISCOVER~\cite{du2024discover} extends it to open-form PDEs.
GenSR~\cite{li2026gensr} searches a learned latent space of equations instead of the discrete equation space.
Xu et al.~\cite{xu2026graphed} represent equations as directed graphs to discover constitutive laws of solids from experiments.
Other studies add physical constraints~\cite{xu2023pic,luo2023physicsguided,du2024rdiscover}, such as dimensional and directional constraints~\cite{cao2022physicalpde}, or combine evolutionary neural networks with physics-informed learning for noisy data~\cite{zhou2024noisypde}.

LLMs bring mathematical knowledge to equation discovery.
LLM4ED~\cite{du2024llm4ed} and EqGPT~\cite{xu2025eqgpt} use LLMs to generate candidate equations, while LLM-SR~\cite{shojaee2025llmsr} evolves equations written as programs.
SR-LLM~\cite{guo2025srllm} retrieves accumulated knowledge, and LaSR~\cite{grayeli2024lasr} guides evolutionary search with natural-language concepts drawn from high-performing hypotheses.
Benchmarks such as SRBench~\cite{lacava2021srbench} and LLM-SRBench~\cite{shojaee2025llmsrbench} compare these methods across problems and show how diverse they are.
These methods complement one another, but each has its own representation, search procedure and data interface.
Some strategies work across methods: DASS~\cite{hu2026dass} guides different methods with an auxiliary search space of reliable terms, and PostDiag~\cite{li2026postdiag} diagnoses and repairs candidates independently of the search.
LegoSR brings seven methods from four families into one framework~(\cref{sec:framework}).
Every method takes the same task and Sketch and returns candidates in Equation IR, so results can be compared directly and reused in later searches.

\subsection{Agentic equation discovery methods}
\label{sec:related_agentic}

LLM agents now take part in candidate generation, search control and experimental refinement.
FunSearch~\cite{romeraparedes2024funsearch} combines language-model-generated programs with systematic evaluation and evolutionary search.
In equation discovery, LLM4PD~\cite{luo2025llm4pd} discovers PDEs from coarse to fine: hierarchical prompting and a review agent identify relevant terms, and a numerical PDE solver then refines their coefficients.
Deliberate Evolution~\cite{pang2026deliberate} separates expression generation from search control and guides generation with structural diagnostics and reflective memory.
SR-Scientist~\cite{xia2026srscientist} gives an agent code tools for data analysis and equation evaluation and lets it refine equations from feedback over long horizons.
DoLQ~\cite{song2026dolq} discovers ODEs with a multi-agent architecture: one agent proposes candidates, a parameter optimizer refines them and a scientist agent evaluates them.

These agents guide the search with intermediate results, each within its own discovery architecture.
KeplerAgent~\cite{yang2026kepleragent} extracts physical structure such as symmetries and uses it to configure PySINDy~\cite{kaptanoglu2022pysindy} or PySR~\cite{cranmer2023pysr}, and LLM-PySR~\cite{xie2026llmpysr} lets LLM agents plan and adjust a PySR search.
In both, the agent configures one or two existing methods, each with its own interface and output.
LegoSR Agent (\cref{sec:agent}) instead coordinates all seven methods through one framework: it selects and configures them, switches between them based on feedback and passes the equations, derivatives and neural surrogates of one search to the next.
When searches fail, it can also build a Sketch from the terms they agree on, through the same interface a researcher uses to add prior knowledge.
Its Experience Library stores lessons about methods, not equations: which derivative estimators, discovery methods and settings worked under which data conditions.
These lessons transfer across equation families.

\section{LegoSR: a composable framework for equation discovery}
\label{sec:framework}

\begin{figure}[!tbp]
\centering
\includegraphics[width=\textwidth]{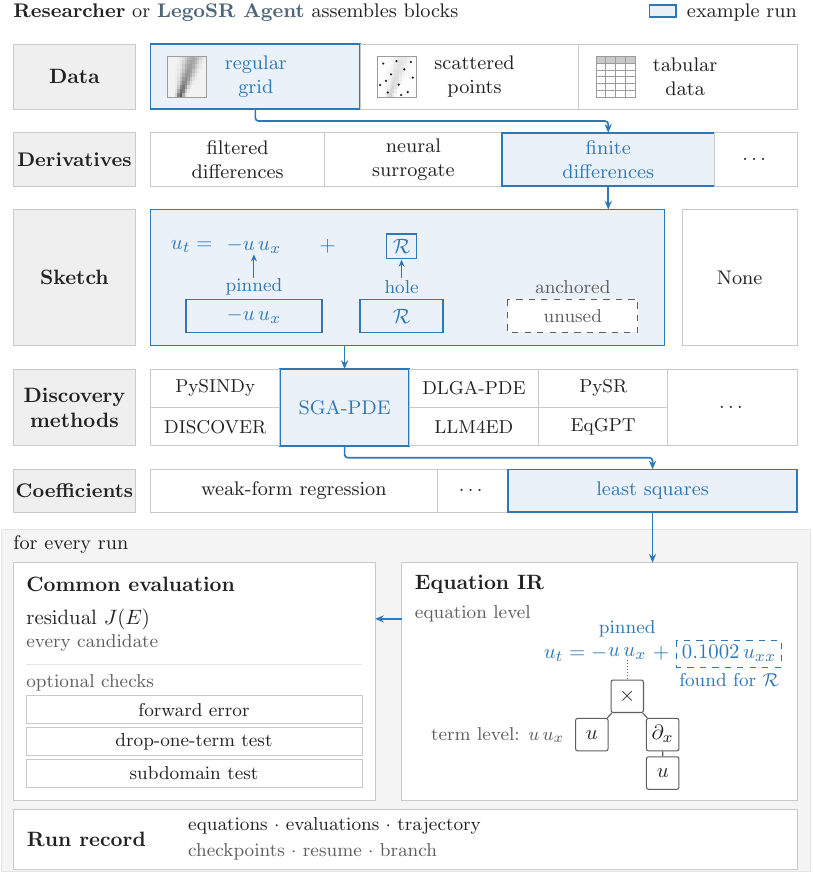}
\caption{Composable equation discovery in LegoSR. Blue blocks and arrows trace an example run for the Burgers equation. In the Sketch, the pinned term $-u\,u_x$ and the hole $\mathcal{R}$ assemble into the prior equation; the anchored component is unused. Pinning fixes a term and its coefficient, whereas anchoring retains a term and fits its coefficient. ``None'' omits the entire Sketch. SGA-PDE searches for the expression in $\mathcal{R}$, yielding $0.1002\,u_{xx}$ after coefficient fitting. Equation IR represents the complete equation and its constituent terms; common evaluation and the run record support comparison and reuse across searches. Dots mark further options.}
\label{fig:architecture}
\end{figure}

LegoSR organizes equation discovery into a unified but flexible workflow of task preparation, equation search, evaluation and experiment management (\cref{fig:architecture}).
It standardizes how these parts exchange data and results: tasks and candidates use common representations (\cref{sec:framework-representation}), discovery methods share computational modules (\cref{sec:framework-solvers}) and every run is evaluated and recorded uniformly (\cref{sec:framework-evaluation}).
Each method keeps its own search strategy and optimization objective, so it can run alone or within a larger discovery process.

\subsection{Representation of complete and partially known equations}
\label{sec:framework-representation}

A discovery task specifies the data and can include prior knowledge as a Sketch.
LegoSR supports regular grids, scattered points and tabular data.
To exchange equations across methods, LegoSR uses a common Equation Intermediate Representation (Equation IR).
Its term level describes expressions built from variables, derivatives, operators and parameters, while its equation level defines the structure, coefficients and constraints.
Equation IR supports algebraic relations, evolution equations and homogeneous relations.
An evolution PDE, for example, can be written as
\begin{equation}
u_t = \sum_{k=1}^{K} \xi_k \phi_k,
\label{eq:ir}
\end{equation}
where $\phi_k$ is a term involving the field, coordinates and derivatives, such as $u\,u_x$ or $u_{xx}$, and $\xi_k$ is its coefficient.
Terms may also contain nonlinear operations and internal parameters.
Discovery methods still search in their native forms, such as expression trees, symbol sequences or candidate-term libraries, and convert their results into Equation IR.

Equation IR expressions compile into Python functions for computation, fitting and evaluation, and discovered expressions can be reused as candidate terms or prior knowledge in later searches~\cite{chen2022knowledgedata}.

A Sketch represents a partially known equation: it specifies the known structure and leaves an expression to be discovered.
For example, if the Burgers advection term and its coefficient are known, we can write
\begin{equation}
u_t = -u\,u_x + \mathcal{R},
\label{eq:sketch}
\end{equation}
where $\mathcal{R}$ is a hole, an unknown expression that may contain one or more terms.
A Sketch can include a known term in two ways.
Pinning fixes both the term and its coefficient: in \cref{eq:sketch}, the term $u\,u_x$ is pinned with coefficient $-1$.
Anchoring keeps the term in every candidate equation but fits its coefficient to the data, together with the terms in $\mathcal{R}$.
Pinning suits a term whose coefficient is known from theory; anchoring suits a term whose form is known but whose coefficient is not, such as a term found by several earlier searches.
The hole can also be constrained by the number of terms, variables, operators and derivative orders.

LegoSR converts a Sketch into a search task for each method and reconstructs the complete equation from the result.
In this example, moving the pinned term to the left gives the target $u_t + u\,u_x$, which the method fits with $\mathcal{R}$.
Different methods can therefore use the same prior knowledge without following the same search procedure.

\subsection{Integrating discovery methods through reusable computational modules}
\label{sec:framework-solvers}

Composable discovery methods let LegoSR Agent and researchers switch between them on the same task and reuse the results of one search in another.
Making them composable requires preserving their distinct search mechanisms while extracting reusable computational modules from implementations tailored to specific experiments.
LegoSR integrates seven representative methods from sparse regression, evolutionary search, reinforcement learning and language models (\cref{tab:solvers}).
Many of their public implementations tie preprocessing, derivative computation, search and evaluation to dedicated scripts.
LegoSR does not just wrap these implementations behind a common interface: it reorganizes their procedures into configurable pipelines and extends several methods to additional problem types.

\begin{table}[!t]
\centering
\caption{Equation discovery methods integrated in LegoSR and the main characteristics of their original designs.}
\label{tab:solvers}
\footnotesize
\begin{tabularx}{\textwidth}{@{}>{\raggedright}p{2.8cm} >{\raggedright}p{2.7cm} >{\raggedright\arraybackslash}X@{}}
\toprule
Method & Approach & Key characteristics \\
\midrule
SINDy/\allowbreak PySINDy~\cite{brunton2016sindy,rudy2017pdefind,kaptanoglu2022pysindy}
& Sparse regression
& Sparse selection from a user-defined library of candidate terms \\
\midrule
DLGA-PDE~\cite{xu2020dlga}
& Neural network and genetic algorithm
& Neural-network derivative estimation and evolutionary construction of candidate terms \\
\midrule
SGA-PDE~\cite{chen2022sgapde}
& Genetic programming
& Open-form discovery of compound and fractional terms using expression trees \\
\midrule
PySR~\cite{cranmer2023pysr}
& Evolutionary symbolic regression
& Joint search over expression structures and numerical constants, balancing accuracy and complexity \\
\midrule
DISCOVER~\cite{du2024discover}
& Reinforcement learning
& Symbol-by-symbol equation generation with constraints on the search space \\
\midrule
LLM4ED~\cite{du2024llm4ed}
& Large language model
& Generation and refinement of candidate equations using prompts and evaluation feedback \\
\midrule
EqGPT~\cite{xu2025eqgpt}
& Pretrained generative model
& Equation generation using mathematical knowledge learned during pretraining \\
\bottomrule
\end{tabularx}
\end{table}

These pipelines draw on shared modules for derivative computation, coefficient optimization by least squares or weak-form regression, data diagnostics, evaluation and visualization.
Derivatives come from finite differences, from finite differences after low-pass or Savitzky--Golay filtering, or from a neural surrogate fitted to the data.
These modules also work outside a discovery pipeline: researchers can inspect derivative quality before a search, or fit and evaluate a candidate equation without rerunning the search.
Keeping them out of method-specific search code also makes existing methods easier to extend to new problems.

A new discovery method needs task adaptation, an execution interface and Equation IR conversion, and then reuses LegoSR's data processing, evaluation, visualization and experiment management.
New derivative estimators, coefficient optimizers and evaluation methods can also be added as shared modules.
Once registered, these capabilities become tools of LegoSR Agent.

\subsection{Evaluation and traceability of discovery results}
\label{sec:framework-evaluation}

Native scores, such as information criteria or rewards, differ across methods, so LegoSR keeps them and adds a common evaluation with the same data processing and derivative estimation for every candidate.
For each candidate $E$, written as in \cref{eq:ir}, LegoSR refits the coefficients by least squares, subject to existing constraints, and reports the normalized residual
\begin{equation}
J(E) = \frac{\lVert u_t - \sum_k \hat{\xi}_k \phi_k \rVert_2^2}{\lVert u_t - \bar{u}_t \rVert_2^2},
\label{eq:residual}
\end{equation}
where $\hat{\xi}_k$ are the refitted coefficients, $\bar{u}_t$ is the mean of $u_t$ and the norms are taken over all data points; $J = 0$ for an exact fit.
For ODEs and evolution PDEs, LegoSR also integrates $E$ from the first observed time and reports the forward error, the relative $L_2$ difference between predicted and observed trajectories or fields.
A drop-one-term test refits $E$ without each term in turn and reports how much $J$ rises.
A subdomain test refits $E$ on four parts of the domain, split along a spatial coordinate, and reports the coefficient spread: the relative standard deviation of each coefficient across the four parts, averaged over the terms.

For each run, LegoSR records its configuration, search trajectory, candidate equations and evaluation results.
It supports checkpoints, resuming interrupted runs, branching from earlier search states, batch experiments and scheduling on CPUs and GPUs.
These records let researchers inspect earlier searches and continue or extend experiments without starting over.
Because every method returns the same evidence, LegoSR Agent can compare the results and use them to plan later searches (\cref{sec:agent}).

\section{LegoSR Agent: adaptive discovery with cumulative experience}
\label{sec:agent}

\begin{figure}[t]
\centering
\includegraphics[width=\textwidth]{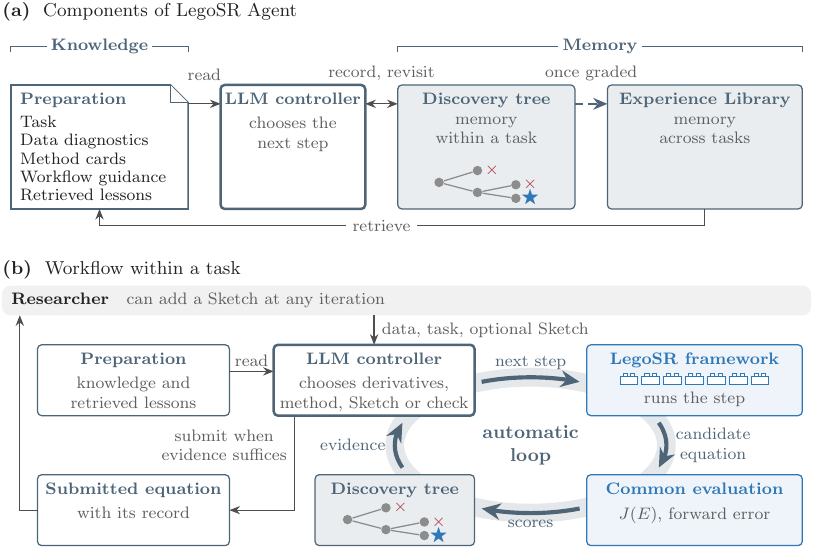}
\caption{LegoSR Agent within a task. (a) Components: knowledge in the preparation, the LLM controller and memory at two time scales (the discovery tree within a task, the Experience Library across tasks). A graded task updates the library (dashed arrow; \cref{fig:agent-workflow}). (b) Workflow: the controller reads the preparation, runs an automatic loop on the LegoSR framework (bold arrows) and submits an equation when the evidence suffices; a researcher can add a Sketch at any iteration. Blue marks the LegoSR framework, slate tint marks memory and grey marks parts outside the agent.}
\label{fig:agent-architecture}
\end{figure}

LegoSR Agent turns the records, evaluations and discovery method interfaces of \cref{sec:framework} into decisions about how discovery should proceed.
It combines knowledge of the discovery methods with experimental memory to guide choices that a generic agent must otherwise infer from tool descriptions.
The agent consists of knowledge, memory and an LLM controller (\cref{fig:agent-architecture}a).
A preparation at the start of each task supplies knowledge of the methods and lessons retrieved for similar data (\cref{sec:agent-knowledge}).
The agent then improves at two time scales, each with its own memory: within a task, a discovery tree retains every step as each iteration narrows the hypothesis (\cref{sec:agent-loop}); across tasks, the Experience Library stores lessons from each graded task (\cref{sec:agent-experience}).

\subsection{Retrieving knowledge to guide discovery decisions}
\label{sec:agent-knowledge}

The preparation gathers the knowledge the controller needs for its first decisions.
The task specification identifies the target and the acceptable answer form.
Method cards summarize each discovery method's strengths, limitations and settings tested in publications and earlier experiments; the controller reads a method's full card when choosing it.
A method registered in LegoSR becomes available to the agent at once, because its card is generated from its registration (\cref{sec:framework-solvers}).
LegoSR computes data diagnostics: grid checks, a noise estimate and estimated errors of the first and second derivatives.
Workflow guidance describes the discovery process of \cref{sec:agent-loop}.

The preparation also includes lessons from past tasks.
They are retrieved by matching the data diagnostics with conditions stored in the Experience Library (\cref{sec:agent-experience}).
The controller thus starts each task knowing which data processing, discovery methods and settings worked or failed on similar data.

\subsection{Refining equation hypotheses through numerical feedback}
\label{sec:agent-loop}

The controller works in a ReAct loop~\cite{yao2023react}, which alternates reasoning and acting (\cref{fig:agent-architecture}b).
In each iteration, it reasons over the evidence in the discovery tree, its knowledge and the remaining budget to choose the next step.
It then acts: LegoSR runs the step and returns derived fields or candidate equations scored by the common evaluation; these results enter the discovery tree as evidence for the next iteration.
In the loop of \cref{fig:agent-architecture}b, the controller chooses among four kinds of step: \textit{derivatives}, \textit{method}, \textit{Sketch} and \textit{check}.
They use, respectively, the derivatives and coefficients stages, the discovery methods stage, the Sketch stage and the common evaluation of \cref{fig:architecture}.

In a \textit{derivatives} step, the controller chooses how to process the data and estimate derivatives: by finite differences, with or without filtering, or through a neural surrogate.
The noisy-data case in \cref{sec:prior_guided} illustrates why this choice precedes symbolic search.
In a \textit{method} step, it selects a discovery method and configures its term library, thresholds and search settings.
Method cards guide changes when progress stalls, as in the switch from PySINDy to SGA-PDE to DLGA-PDE in \cref{sec:domain_guided}.
In a \textit{Sketch} step, it constructs or revises a Sketch, which turns failed searches into new hypotheses: terms on which the searches agree are anchored, and disputed terms are left as holes.
In \cref{sec:prior_guided}, the agent builds a Sketch at 10\% noise, and a researcher writes the same Sketch at 25\% noise.
In a \textit{check} step, it chooses how to examine the candidates: remove small terms and refit with another module, seek confirmation from an independent discovery method, test whether each term is needed or integrate the equation and compare its dynamics with observations (\cref{sec:quickstart}).

LegoSR also reports the progress of each search, so the controller can decide whether to continue it, adjust its configuration or switch methods.
Each step enters the discovery tree under the step it builds on, with its candidate equations, evaluations and rationale.
Because the tree retains every step of the task, the controller can revisit earlier branches to continue promising searches or change direction without repeating unproductive attempts.
Over successive iterations, the controller rejects some terms, anchors others and leaves the rest as holes until it finds and checks the equation (\cref{sec:prior_guided}).
When the evidence suffices, the controller submits the selected equation with its supporting record.

The controller's actions operate on LegoSR's shared objects: observed and derived fields, Sketches and candidates in Equation IR.
These objects support comparison and reuse across discovery methods; for example, a neural surrogate from an unsuccessful search supplies derivatives for later steps in \cref{sec:prior_guided}.
Both the agent and a researcher express structural hypotheses through the same Sketch interface, so a researcher can step in at any iteration.
LegoSR measures and records; the controller decides.

\subsection{Transferring discovery experience across tasks}
\label{sec:agent-experience}

An experienced researcher starts a new equation discovery problem with lessons from earlier ones: which data processing, discovery method and setting suit which data.
Without memory across tasks, an agent spends its search budget repeating earlier trials.
The Experience Library gives LegoSR Agent this cumulative experience; it is the agent's memory across tasks (\cref{fig:agent-architecture}a).
The library starts from earlier experiments and grows with every task graded against a reference equation.

The difficulty is that experience must transfer to equations the agent has never seen.
A lesson that recorded the terms of an earlier equation would help only on equations with the same terms.
The Experience Library therefore stores lessons about choices under data conditions, such as high noise on a one-dimensional grid, not about equations.
Data conditions are measured on the data before the search begins, so a lesson applies to every new task whose data meet its condition.

\begin{figure}[t]
\centering
\includegraphics[width=\textwidth]{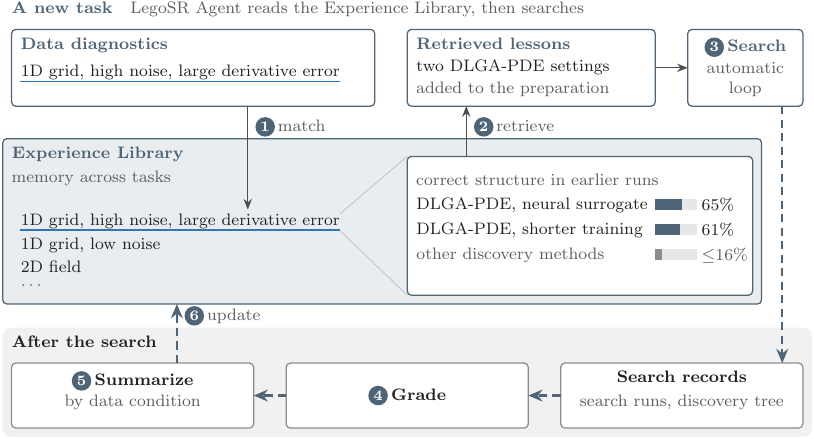}
\caption{How LegoSR Agent improves across tasks. Numbers give the order of the steps. A new task matches its data diagnostics with conditions in the Experience Library and adds the retrieved lessons to its preparation (\cref{fig:agent-architecture}a) before the search. After the search, its records are graded and summarized by data condition to update the library (grey band, dashed arrows). In the matched entry, each percentage is the fraction of earlier runs with that setting that recovered the correct structure.}
\label{fig:agent-workflow}
\end{figure}

Each lesson links a data condition to a choice of data processing, derivative estimation and discovery method with its settings, and records how that choice fared in earlier runs.
\Cref{fig:agent-workflow} shows an entry of the library.
At the start of a new task, the agent matches its data diagnostics with the stored conditions, and the retrieved lessons enter its preparation (steps 1 and 2 in \cref{fig:agent-workflow}).
After the task, its search records are graded and the outcomes update the library (steps 4--6).
Across tasks, accumulated lessons change the guidance for later decisions, as the discovery tree does for later steps within a task.

Lessons transfer across equation families: in our experiments, each task uses only lessons from other families.
In \cref{sec:prior_guided}, after the first searches fail, the agent follows a retrieved lesson and takes derivatives from a neural surrogate.
\Cref{sec:experiments} compares LegoSR Agent with a generic agent that has the same model, tools and budget but no method cards, workflow guidance or Experience Library.

\section{Experiments}
\label{sec:experiments}

We test whether LegoSR Agent, deciding on its own, reaches the coverage of expert-tuned methods and how much its method cards and experience improve coverage and efficiency over a generic agent.
As a strong reference, we construct the Virtual Best Solver (VBS), which combines the successful recoveries of up to seven applicable expert-tuned methods, and compare it with a generic agent and LegoSR Agent across tasks and noise levels.

\subsection{Experimental Setup}
\label{sec:exp-setup}

\subsubsection{Benchmark problems}

The benchmarks contain 134 algebraic SR problems, 40 ODE problems and 25 PDE problems.

The algebraic SR benchmark consists of 120 Feynman equations and 14 equations from seven two-dimensional Strogatz systems.
The Feynman equations and variable ranges come from the AI Feynman database~\cite{udrescu2020aifeynman}.
We independently sample each variable uniformly within its range, generating 10,000 training and 2,500 test samples per problem.
For the Strogatz systems from SRBench~\cite{lacava2021srbench}, we regenerate trajectories through numerical integration, compute the targets from the ground-truth equations and apply a 75/25 training--test split.

The ODE benchmark contains 40 target equations from 19 dynamical systems, including the Lorenz, Duffing, Van der Pol, SIR and FitzHugh--Nagumo systems.
Unlike algebraic SR, which uses ground-truth right-hand sides as regression targets, ODE discovery requires derivative estimation from observed trajectories.
The PDE benchmark consists of 25 one- and two-dimensional problems, including the Burgers, Korteweg--de Vries (KdV), Allen--Cahn, Kuramoto--Sivashinsky, Klein--Gordon, wave, beam, nonlinear Schr\"odinger, reaction--diffusion and variable-coefficient diffusion equations.
The problems are listed in \cref{app:benchmarks}.

\subsubsection{Noise levels and recovery criteria}

We add independent Gaussian noise at three levels (0\%, 1\% and 5\%) to the training data and keep the test data noise-free.
For algebraic SR, noise is added to the target $y$ with standard deviation $\eta\,\mathrm{RMS}(y)$, where $\eta$ denotes the noise level.
For ODEs and PDEs, noise is added to the observed trajectories or fields before derivative estimation, scaled by the standard deviation of each variable.
All approaches use the same noisy data for each problem.

We report recovery coverage, defined as the percentage of problems for which at least one run recovers the target equation.
Each agent runs three times per problem.
For algebraic SR, recovery follows the SRBench symbolic equivalence criterion~\cite{lacava2021srbench}: after rounding to three significant digits and simplification, the predicted expression must not reduce to a constant, and its difference from the ground truth must be constant or their ratio a nonzero constant.
We also require a test $R^2 > 0.5$ to exclude inaccurate expressions.

For ODEs, recovery also requires the predicted right-hand side to agree with the true one within 5\% at sampled points.
For PDEs, recovery requires the exact set of ground-truth terms and a maximum relative coefficient error of at most 5\%:
\begin{equation}
  \epsilon_{\max} = \max_k \frac{|\hat{\xi}_k - \xi_k^{\mathrm{true}}|}{|\xi_k^{\mathrm{true}}|} \leq 0.05,
  \label{eq:coef-error}
\end{equation}
where $\hat{\xi}_k$ and $\xi_k^{\mathrm{true}}$ are the submitted and true coefficients of the term $\phi_k$ in \cref{eq:ir}.
Submitted coefficients are graded without refitting.

\subsubsection{Compared approaches}

\textbf{Generic Agent.}
The generic agent receives a basic task description and has access to all LegoSR tools.

\textbf{LegoSR Agent.}
LegoSR Agent additionally uses method cards, workflow guidance and the Experience Library (\cref{sec:agent}) for autonomous method selection, configuration and multi-stage search.

Both agents use the DeepSeek-V4.1-Flash-1017 model (\texttt{deepseek-flash} in the DeepSeek API) and run independently three times per problem under identical budgets.
Each run has a 60-minute soft limit, a 120-minute hard limit and at most 400 steps.
Responses are limited to 16,384 tokens, with a 600-second timeout per request.
Input data are anonymized by removing problem names, ground-truth equations and original variable names.

\textbf{VBS.}
VBS serves as a standard and highly demanding reference for the evaluated methods.
We collect recommended configurations from original papers, open-source code and author communications, supplemented by effective settings from our experiments.
For SR and ODE tasks, we use these configurations with three seeds per applicable method.
For PDE tasks, we further conduct extensive ground-truth-guided offline tuning, involving approximately 10,000 runs and consuming 80 NVIDIA A100 GPU-hours and 1,000 CPU core-hours on AMD EPYC 7763 processors; we denote this reference VBS$^{\dagger}$.
We retain the best configuration found for each method and problem.
A problem is counted as recovered if any run with these configurations succeeds.

\subsection{Recovery of Algebraic and Ordinary Differential Equations}
\label{sec:exp-overall}

LegoSR Agent achieves coverage comparable to VBS across all noise levels on algebraic SR problems (\cref{tab:sr-results}).
While VBS pools 15 runs from five applicable expert-tuned methods, LegoSR Agent reaches similar coverage with only three agent runs.
Algebraic SR involves no derivative estimation or field preprocessing, so the three approaches perform alike.
The differences appear in differential equation discovery, where these choices matter.

On ODEs, LegoSR Agent outperforms the generic agent at every noise level and stays close to VBS (\cref{tab:sr-results}).
At 5\% noise, LegoSR Agent improves ODE coverage by 5 percentage points over the generic agent.

\begin{table}[!htb]
\centering
\caption{Recovery coverage on 134 algebraic SR and 40 ODE problems. Numbers of recovered problems are shown in parentheses.}
\label{tab:sr-results}
\footnotesize
\begin{tabular}{llrrr}
\toprule
Task & Noise & VBS & Generic Agent & LegoSR Agent \\
\midrule
\multirow{3}{*}{Algebraic SR (134)}
& 0\% & 52.2\% (70) & 50.0\% (67) & 48.5\% (65) \\
& 1\% & 50.7\% (68) & 48.5\% (65) & 50.0\% (67) \\
& 5\% & 48.5\% (65) & 44.0\% (59) & 45.5\% (61) \\
\midrule
\multirow{3}{*}{ODE (40)}
& 0\% & 70.0\% (28) & 72.5\% (29) & 77.5\% (31) \\
& 1\% & 55.0\% (22) & 42.5\% (17) & 50.0\% (20) \\
& 5\% & 42.5\% (17) & 37.5\% (15) & 42.5\% (17) \\
\bottomrule
\end{tabular}
\end{table}

\subsection{PDE Discovery}

\subsubsection{Recovery under Noise}

LegoSR Agent recovers all 25 PDE problems without noise and maintains 84\% recovery coverage at both 1\% and 5\% noise (\cref{tab:de-results}).
Its advantage over the generic agent widens as noise increases.

\begin{table}[!htb]
\centering
\caption{Recovery coverage on 25 PDE problems. Numbers of recovered problems are shown in parentheses. VBS$^{\dagger}$ includes extensive ground-truth-guided offline tuning.}
\label{tab:de-results}
\footnotesize
\begin{tabular}{lrrr}
\toprule
Noise & VBS$^{\dagger}$ & Generic Agent & LegoSR Agent \\
\midrule
0\% & 100.0\% (25) & 92.0\% (23) & 100.0\% (25) \\
1\% & 80.0\% (20) & 60.0\% (15) & 84.0\% (21) \\
5\% & 68.0\% (17) & 56.0\% (14) & 84.0\% (21) \\
\bottomrule
\end{tabular}
\end{table}

At 1\% and 5\% noise, LegoSR Agent recovers one and four more problems than VBS$^{\dagger}$, respectively.
VBS$^{\dagger}$ counts a problem only when a single method run succeeds, whereas LegoSR Agent combines partial findings across methods.
The search traces show that failed method runs still yield useful structures, even with extra terms or inaccurate coefficients.
LegoSR Agent takes derivatives from neural surrogates, combines these structures into a Sketch, tests them on the data and refits their coefficients by weak-form regression.
\Cref{sec:prior_guided} follows this process step by step on the Chafee--Infante equation.

\subsubsection{Computational Efficiency}

\begin{figure}[!htb]
\centering
\includegraphics[width=\textwidth]{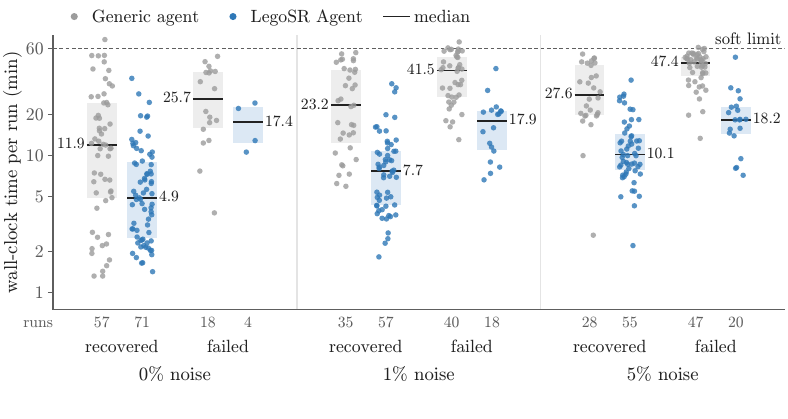}
\caption{Wall-clock time of the generic agent and LegoSR Agent on the 25 PDE problems, three runs per problem, grouped by recovery outcome. Points represent runs, bars mark medians, shaded boxes show interquartile ranges, and numbers indicate run counts.}
\label{fig:efficiency}
\end{figure}

LegoSR Agent is substantially faster than the generic agent on PDE tasks: over all runs, its median wall-clock time is 63--73\% shorter across the three noise levels.
\Cref{fig:efficiency} shows that the efficiency gain holds for both successful and failed runs.
At 5\% noise, the median time falls from 27.6 to 10.1 minutes for successful runs and from 47.4 to 18.2 minutes for failed runs, while LegoSR Agent succeeds in 55 of 75 runs against 28 for the generic agent.
Guided by its knowledge and experience, LegoSR Agent uses intermediate feedback to adjust configurations, switch methods or abandon unpromising searches.

Even a single pass of VBS is slow: running all seven methods once takes about 48 minutes per problem on average.
Building the PDE VBS$^{\dagger}$ further required about 10,000 offline runs, 80 GPU-hours, 1,000 CPU core-hours and substantial manual tuning (\cref{sec:exp-setup}).
With default or recommended configurations, the seven methods recover only 40 of the 75 instances, each a PDE problem at one noise level; this tuning raises the number to 62.
VBS$^{\dagger}$ is therefore a demanding reference: it keeps the best expert-tuned runs, tuned with knowledge of the target equations.
LegoSR Agent still exceeds its coverage at 1\% and 5\% noise without access to the ground-truth equations.

\subsubsection{Ablation Study}
\label{sec:exp-ablation}

We ablate LegoSR Agent in two steps.
The generic agent lacks method cards, workflow guidance and the Experience Library; LegoSR Agent w/o Experience Library adds back the first two.
Each configuration runs three times on every PDE problem at each noise level, giving 225 runs; unlike the coverage in \cref{tab:de-results}, \cref{tab:ablation} counts every successful run.
Method cards and workflow guidance give the agent an effective base for the search: they raise the successful runs from 120 to 161.
The Experience Library adds 22 more, and its gain grows with noise, from 3 runs without noise to 7 and 12 runs at 1\% and 5\% noise.

\begin{table}[!htb]
\centering
\caption{Successful runs in the ablation of the Experience Library on 25 PDE problems. Each configuration has 75 runs per noise level and 225 in total.}
\label{tab:ablation}
\footnotesize
\begin{tabular}{lrrrr}
\toprule
Configuration & 0\% & 1\% & 5\% & Total \\
\midrule
Generic Agent & 57 & 35 & 28 & 120 \\
LegoSR Agent w/o Experience Library & 68 & 50 & 43 & 161 \\
LegoSR Agent & 71 & 57 & 55 & 183 \\
\bottomrule
\end{tabular}
\end{table}

\begin{figure}[!htb]
\centering
\includegraphics[width=0.62\textwidth]{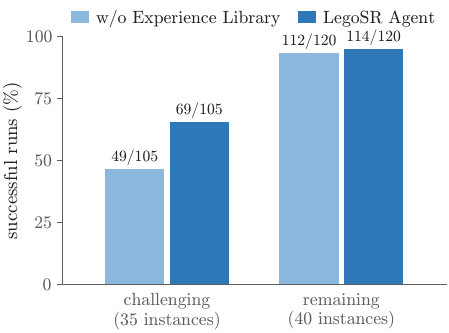}
\caption{Successful runs of LegoSR Agent with and without the Experience Library on challenging and remaining instances, three runs per instance.}
\label{fig:ablation-difficulty}
\end{figure}

We label an instance challenging if none of the seven methods recovers it with its default or recommended configuration, that is, before the ground-truth-guided tuning of VBS$^{\dagger}$.
Of the 75 instances, 35 are challenging.
The gain of the Experience Library concentrates on these instances (\cref{fig:ablation-difficulty}).
On the remaining 40 instances, LegoSR Agent succeeds in almost every run with or without the library.
On the challenging instances, the library raises the successful runs from 49 to 69 of 105.
Paired by problem, noise level and random seed, it turns 26 failed runs into successes and 6 successful runs into failures (two-sided sign test, $p = 0.0005$).
In these instances the choice of data processing, derivative estimation and settings decides the outcome, and lessons from other equation families guide this choice (\cref{sec:agent-experience}).

Both components also shorten the search.
Method cards and workflow guidance reduce the median wall-clock time per run from 14.3, 32.6 and 43.9 minutes to 7.7, 15.3 and 23.0 minutes at 0\%, 1\% and 5\% noise.
The Experience Library reduces it further to 5.3, 8.9 and 12.1 minutes.

\section{Case studies of adaptive equation discovery}
\label{sec:case-study}

The three case studies below increase in difficulty.
\Cref{sec:quickstart} shows that a few lines of code or
a single command to LegoSR Agent recovers the noise-free Burgers equation.
In \cref{sec:domain_guided}, SGA-PDE recovers
the correct terms of the KdV
equation with 1\% noise, but noisy third-order derivatives
bias their coefficients; LegoSR Agent traces the bias
with its method cards and switches methods.
In \cref{sec:prior_guided}, no single discovery method
reliably finds the
structure of the Chafee--Infante equation with 10\% and
25\% noise, so LegoSR Agent builds a candidate structure
from failed searches, or a researcher supplies one.
The role of the agent thus grows from running a discovery method
to choosing one and then to building hypotheses.

\subsection{Quick start: from data to equations}
\label{sec:quickstart}

To illustrate LegoSR's end-to-end workflow, we consider
the one-dimensional viscous Burgers equation,
\begin{equation}
    u_t = -u\,u_x + 0.1\,u_{xx}.
    \label{eq:burgers_case}
\end{equation}
The noise-free dataset from SGA-PDE~\cite{chen2022sgapde} samples $u(x,t)$ on a $256 \times 201$ grid with $x \in [-8,8)$, $t \in [0,10]$ and periodic boundaries.

\subsubsection{Manual workflow}

LegoSR provides a unified Python interface for data loading, method configuration, equation discovery and result analysis. It can be installed via \verb|pip install "sail-kd[agent]"|. The following example runs SGA-PDE:

\par\nopagebreak\noindent\begin{minipage}{\linewidth}
\begin{lstlisting}[language=Python]
import kd

data = kd.load("burgers")
model = kd.Model(algorithm="sga",
                 generations=30,
                 population=20, seed=0)
model.fit(data)

model.report("out/burgers/")
\end{lstlisting}
\end{minipage}\par

The recovered equation is
$u_t = -1.000\,u\,u_x + 0.100\,u_{xx}$,
closely matching the ground truth.
Because the discovery methods share one interface and equation representation, setting the \texttt{algorithm} parameter switches to another method, such as PySINDy, without changing the data or analysis workflow.

\begin{figure}[!htb]
    \centering
    \includegraphics[width=\textwidth]{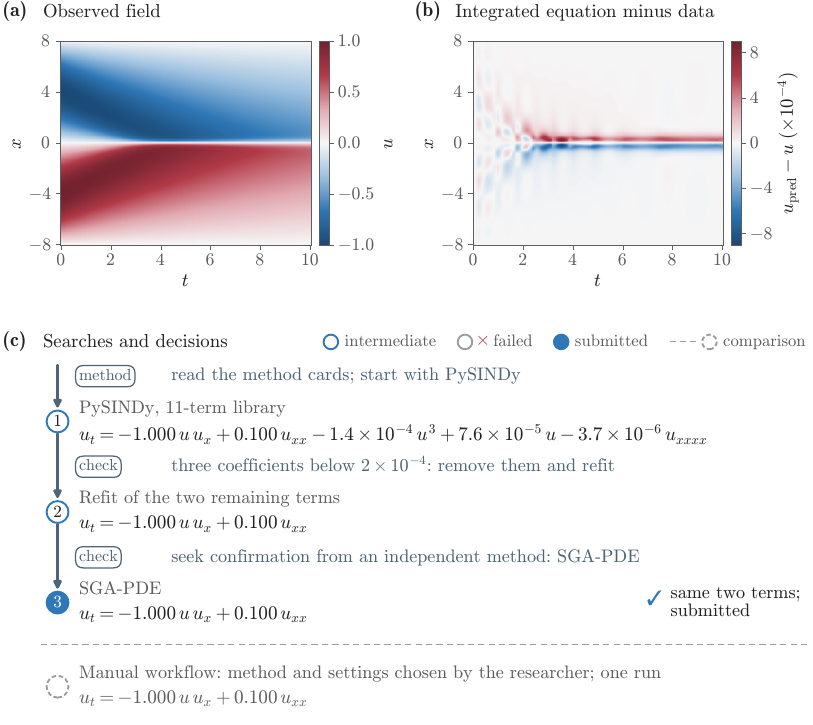}
    \caption{Burgers equation recovered with LegoSR and LegoSR Agent.
    (a) Observed field.
    (b) Solution of the recovered equation, integrated from $t=0$, minus the data;
    (a) and (b) are from the output of \texttt{model.report()}.
    (c) Searches and decisions of LegoSR Agent in one of its three runs; below the dashed line, the single SGA-PDE run of the manual workflow.}
    \label{fig:quickstart}
\end{figure}

LegoSR also generates diagnostic reports for inspecting
the discovered equations. \Cref{fig:quickstart}a
shows the observed field, while \cref{fig:quickstart}b shows the error of the recovered equation integrated forward in time.
The relative $L_2$ error is $1.8 \times 10^{-4}$,
with the largest discrepancies near steep gradients.

LegoSR Agent takes over the manual workflow's decisions:
which method to run, how to configure it, how to evaluate
candidates and what to do next.

\subsubsection{Autonomous discovery with LegoSR Agent}
Researchers can hand the task to LegoSR Agent
with one command:

\par\nopagebreak\noindent\begin{minipage}{\linewidth}
\begin{lstlisting}[language=bash]
kd-agent run --data burgers.npz "Find the equation governing this data."
\end{lstlisting}
\end{minipage}\par

The agent receives the first 161 time frames ($t \in [0, 8]$);
the remaining frames are held out for evaluation.
The agent first reads the method cards and available
configurations, then applies PySINDy with an expanded library to obtain an initial five-term
equation. After pruning terms with negligible coefficients, it refits the remaining two terms. The agent then runs SGA-PDE for an independent search,
recovering the same two-term structure before submitting
the final equation (\cref{fig:quickstart}c).

LegoSR Agent recovers the governing equation in all three independent runs. Tool calls, intermediate candidates and results are recorded, so researchers can trace the discovery process.
The appendices include a step-by-step walkthrough and GUI examples.

\subsection{Knowledge-guided method selection}
\label{sec:domain_guided}

We consider the KdV equation with 1\% observational noise:
\begin{equation}
    u_t = -u\,u_x - 0.0025\,u_{xxx}.
    \label{eq:kdv_case}
\end{equation}
The equation combines nonlinear convection with
third-order dispersion. The small dispersion coefficient
and the sensitivity of high-order derivatives to noise
make accurate recovery challenging.

\begin{figure}[tbp]
    \centering
    \includegraphics[width=\textwidth]{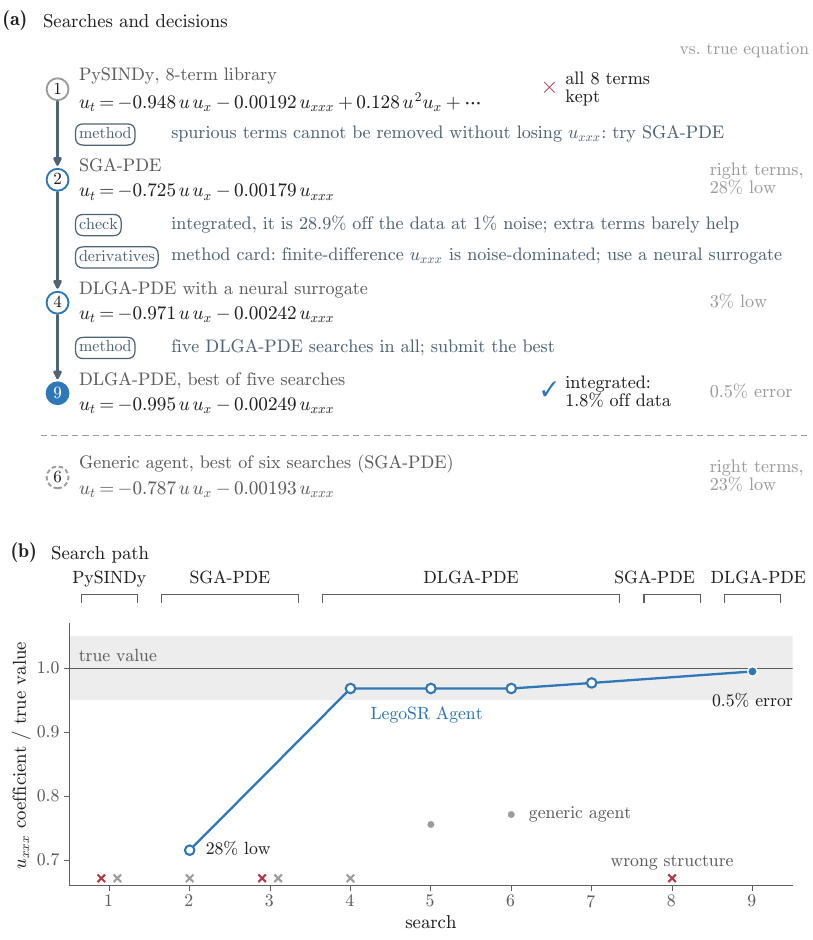}
    \caption{Search for the KdV equation with 1\% noise.
    (a) Equations from selected searches of LegoSR Agent and the decisions between them, with the best search of the generic agent below the dashed line; numbers refer to the searches in (b), and the right column compares each equation with the true one; symbols as in \cref{fig:quickstart}c.
    (b) $u_{xxx}$ coefficient relative to the true value after each search of LegoSR Agent (blue) and the generic agent (gray); crosses mark wrong structures (red: LegoSR Agent), and the band marks $\pm5\%$; method labels refer to the searches of LegoSR Agent.}
    \label{fig:kdv_search}
\end{figure}

We compare LegoSR Agent with the generic agent of
\cref{sec:exp-setup}. As in \cref{sec:quickstart}, no discovery method is specified, and each agent receives the first 161 frames ($t \in [0, 0.8]$); the rest are held out.

LegoSR Agent first applies PySINDy (\cref{fig:kdv_search}), but sparse
regression cannot remove the spurious terms without
also removing the small dispersion term. The agent
then switches to SGA-PDE, which returns the two-term
equation $u_t=-0.725\,u\,u_x-0.00179\,u_{xxx}$.
This structure is physically plausible, since the
steepening and splitting of the waves in the data
point to nonlinear convection and dispersion.
However, when the agent integrates this equation from
the first observed snapshot, the solution deviates
from the data by 28.9\% in relative $L_2$ norm, far
above the 1\% noise level. Adding further terms barely
reduces the fitting residual, so the agent attributes
the mismatch to inaccurate coefficients rather than a
missing term. Its method cards note that, at this
noise level, finite-difference estimates of
third-order derivatives are dominated by noise.
The agent therefore switches to DLGA-PDE (search 4 in
\cref{fig:kdv_search}b), which takes derivatives from a
neural surrogate of the field. The best of its five
DLGA-PDE searches gives the final equation
\begin{equation}
    u_t = -0.995\,u\,u_x - 0.00249\,u_{xxx},
    \label{eq:kdv_recovered}
\end{equation}
and its forward error drops to 1.8\%,
close to the 1.6\% obtained with the true equation
on the same data.

\begin{figure}[!tp]
    \centering
    \includegraphics[width=\textwidth]{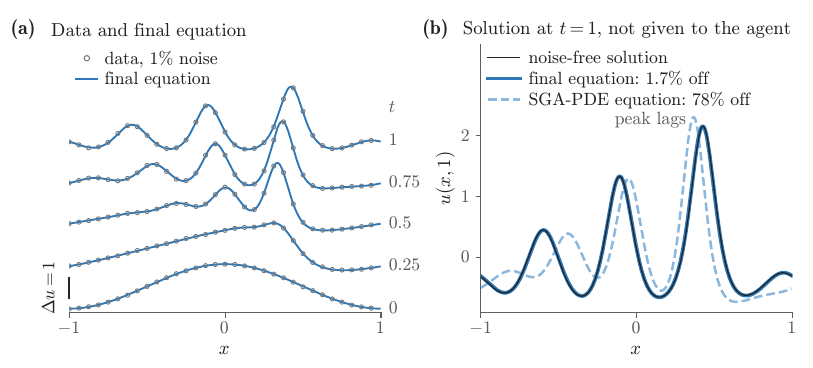}
    \caption{Forward integration of the recovered KdV equations.
    (a) Data and the final equation integrated from $t=0$.
    (b) Profiles at $t=1$, beyond the time window given to the agent, integrated from the noise-free initial state.}
    \label{fig:kdv_check}
\end{figure}

Comparison with the true equation confirms this
diagnosis. The final coefficients are within 0.5\%
of the true values, whereas both SGA-PDE coefficients
are about 28\% too low. The contrast is clearest at
$t=1$, beyond the time window given to the agent.
Integrated from the noise-free initial state, the
SGA-PDE equation deviates from the noise-free profile
by 78\% at this time, while the final equation stays
within 1.7\% (\cref{fig:kdv_check}).
The generic agent repeatedly runs SGA-PDE and
PySR. Its best equation also comes from SGA-PDE and
has the correct terms, but its coefficients are
21\% and 23\% too low (\cref{fig:kdv_search}a).

Both agents reach the correct structure, but only
LegoSR Agent, guided by its method cards, traces the
biased coefficients to noisy third-order derivatives
and corrects them by switching methods.
These switches are direct because all discovery methods share one interface (\cref{sec:framework}): they take the same data and return candidates in Equation IR, all scored by the common evaluation.
Shared modules for derivative computation and coefficient optimization (\cref{sec:framework-solvers}) let the agent change the derivative estimator, the discovery method and the coefficient fit independently (\cref{sec:prior_guided}).
Without such an interface, each switch would require converting data, derivatives and results by hand between separate implementations.
This building-block design is a core contribution of this work.

\subsection{Adaptive and prior-guided discovery}
\label{sec:prior_guided}

We consider the Chafee--Infante equation,
\begin{equation}
    u_t = u_{xx} - u + u^3,
    \label{eq:chafee_case}
\end{equation}
under 10\% and 25\% observational noise.

\begin{figure}[tbp]
    \centering
    \includegraphics[width=\textwidth]{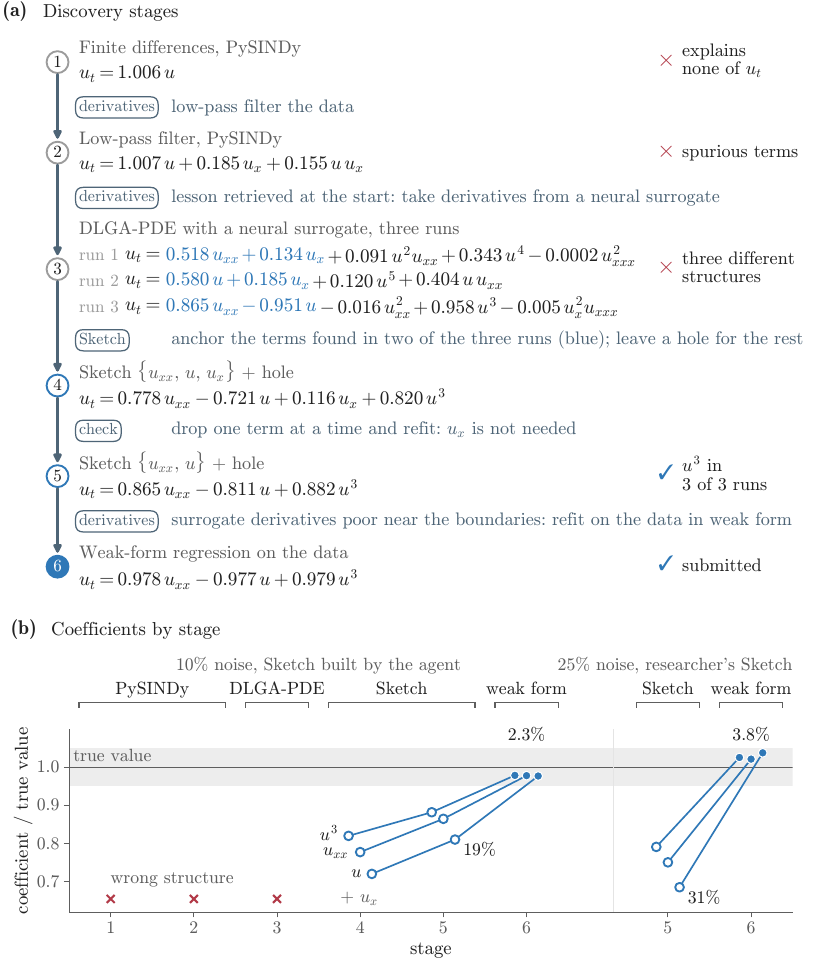}
    \caption{Discovery of the Chafee--Infante equation.
    (a) Discovery stages at 10\% noise; stage 3 lists the three DLGA-PDE runs, with the terms found in at least two runs in blue, and stages 4--6 follow the first run; \cref{fig:chafee_checks}a,b show the evidence behind the last two \textit{derivatives} decisions and the \textit{check} decision; symbols as in \cref{fig:quickstart}c.
    (b) Coefficients relative to the true values after stages 4--6 at 10\% noise and after the Sketch search and the weak-form fit at 25\% noise; crosses mark wrong structures, and the band marks $\pm5\%$.}
    \label{fig:chafee_stages}
\end{figure}

\begin{figure}[tbp]
    \centering
    \includegraphics[width=\textwidth]{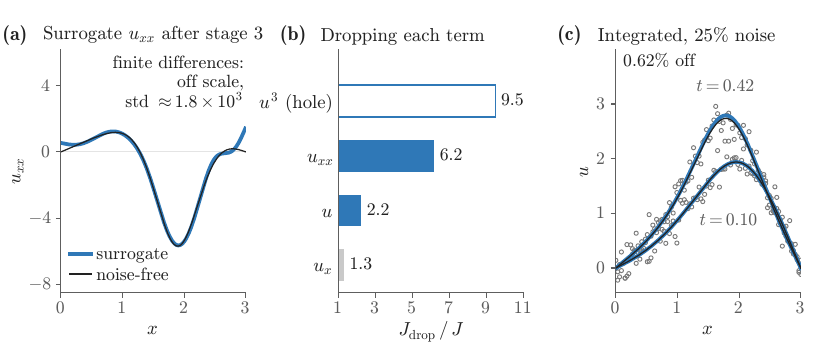}
    \caption{Checks in the Chafee--Infante case.
    (a) $u_{xx}$ at $t=0.26$ from the surrogate of the first stage-3 run and from the noise-free field; finite differences of the noisy data are off scale.
    (b) Normalized residual $J$ after dropping each term of the stage-4 equation, relative to $J$ of the full equation.
    (c) Data with 25\% noise (circles), the recovered equation integrated from the noise-free initial state (blue) and the noise-free field (black).}
    \label{fig:chafee_checks}
\end{figure}

At 10\% noise, the first searches fail (stages 1--3 in \cref{fig:chafee_stages}a).
PySINDy returns wrong equations with or without low-pass filtering, and three DLGA-PDE runs with different seeds give three different structures.
\mbox{However}, each DLGA-PDE run leaves a neural surrogate of the field.
The derivatives of the surrogate are smooth, whereas finite differences of the noisy data are dominated by noise (\cref{fig:chafee_checks}a), so the agent uses the surrogate in later steps, as the retrieved lessons recommend (\cref{fig:agent-workflow}).

Instead of restarting the search, the agent builds on these failed runs.
It writes a Sketch (\cref{sec:framework-representation}) that anchors the terms found in at least two of the three runs, $u_{xx}$, $u$ and $u_x$, and leaves a hole for the nonlinear term, on which the runs disagree.
A search on the surrogate fills the hole with $u^3$ (stage 4).
The agent then tests this equation by dropping one term at a time and refitting the others (\cref{fig:chafee_checks}b).
Dropping $u_x$ increases the normalized residual $J$ of \cref{eq:residual} by only 30\%, so $u_x$ is removed from the anchors.
Dropping $u^3$ increases $J$ by a factor of 9.5, which indicates that a nonlinear term is needed but does not confirm its form.
The agent therefore keeps the hole open and searches again with only $u_{xx}$ and $u$ anchored (stage 5).

In all three runs, this search again fills the hole with $u^3$.
Because the derivatives of the surrogate are least accurate near the domain boundaries (\cref{fig:chafee_checks}a), the agent refits the coefficients by weak-form regression~\cite{messenger2021wsindy} on the observed data (stage 6).
Weak-form regression integrates the equation against smooth test functions and requires no derivatives of the data.
It reduces the maximum coefficient error from 19\% to 2.3\% (\cref{fig:chafee_stages}b), giving
\begin{equation}
    u_t = 0.978\,u_{xx} - 0.977\,u + 0.979\,u^3.
    \label{eq:chafee_recovered_10}
\end{equation}
Integrated forward, it reproduces the noise-free field within 0.56\%.
This workflow combines blocks from different methods within one task: the derivatives come from the surrogate of a DLGA-PDE run, the structure from the Sketch search and the coefficients from weak-form regression (\cref{fig:architecture}).

At 25\% noise, the intermediate results no longer separate the correct terms from spurious ones, so a researcher supplies the Sketch directly:

\par\nopagebreak\noindent\begin{minipage}{\linewidth}
\begin{lstlisting}[language=Python]
sketch = kd.Sketch(
    anchored=["u_xx", "u"],
    holes=[kd.Hole(max_terms=2)]
)
\end{lstlisting}
\end{minipage}\par

In all three runs, the search fills the hole with $u^3$.
Weak-form regression, with a correction for the noise bias in the $u^3$ term, then reduces the maximum coefficient error from 31\% to 3.8\% (\cref{fig:chafee_stages}b), giving
\begin{equation}
    u_t = 1.022\,u_{xx} - 1.038\,u + 1.026\,u^3.
    \label{eq:chafee_recovered_25}
\end{equation}
Integrated forward, it reproduces the noise-free field within 0.62\% (\cref{fig:chafee_checks}c).

The two noise levels illustrate two uses of the same Sketch interface: at 10\% noise, the agent builds the Sketch from failed searches, whereas at 25\% noise, a researcher provides it from domain knowledge.
In both cases, the agent completes the search and refines the coefficients.

\section{Conclusion}
\label{sec:conclusion}

Equations discovered from data serve as governing laws, constitutive relations and closure models, but finding them still depends on experts who choose methods, read intermediate results and decide the next search.
We presented LegoSR, a unified framework for agent-guided equation discovery that integrates seven methods through a common equation representation and evaluation.
Researchers compare candidate equations across methods and check them against the data, and every experiment is traceable and resumable.
Partially known equations constrain the search, so physical knowledge enters the discovery directly.
Building on this framework, researchers delegate the whole discovery process to LegoSR Agent.
Within a task, the agent selects and configures methods, reads intermediate results and uses unsuccessful searches to guide the next steps.
Across tasks, lessons from its graded searches guide later tasks.
By combining partial findings across methods, LegoSR Agent recovers more noisy PDEs than seven methods combined, each tuned with knowledge of the target equations, and it stays close to the expert-configured methods on ODEs.
It also outperforms a generic agent with the same model, tools and budget by up to 28 percentage points in PDE recovery coverage while running substantially faster.
The Experience Library helps most on challenging PDE instances: where no method succeeds with its default or recommended configuration, it raises the share of successful runs from 47\% to 66\%.
In the case studies, knowledge-guided method selection corrects biased KdV coefficients, and LegoSR Agent builds a partially known equation from failed searches to recover the Chafee--Infante equation at 10\% noise.
By connecting individual searches and reusing experience across tasks, LegoSR turns equation discovery from isolated runs into a continuous, adaptive process.
Future work will extend the framework to experimental data, additional discovery methods and broader experience reuse.

\section*{Acknowledgments}
We thank Shuai Lu for contributions to this work.
The authors gratefully acknowledge the High-performance Computing Platform of Eastern Institute of Technology, Ningbo, for providing computational resources for this research.
This work is not related to Xu T. Liu's position at Amazon.

\clearpage
\appendix
\makeatletter
\@addtoreset{table}{section}
\@addtoreset{figure}{section}
\makeatother
\renewcommand{\thesubsection}{\Alph{section}.\arabic{subsection}}

\expandafter\def\expandafter\UrlBreaks\expandafter{\UrlBreaks\do\-}

\section{Tutorial: from installation to equation discovery}
\label[appendix]{app:tutorial}

This appendix walks through a first use of LegoSR.
It installs the package, configures a language model and recovers the Burgers equation of \cref{sec:quickstart} with Python code.
It then checks candidate equations on noisy data and hands the Burgers task to LegoSR Agent.
The online documentation (\href{https://scientific-artificial-intelligence-lab.github.io/kd/}{\nolinkurl{scientific-artificial-intelligence-lab.github.io/kd}}) describes every function used here, and the repository includes Jupyter notebooks with recorded outputs (\href{https://github.com/scientific-artificial-intelligence-lab/KD/tree/trunk/examples/notebooks}{\nolinkurl{examples/notebooks}}).

\subsection{Installation}
\label[subappendix]{app:install}

LegoSR is released as the open-source Python package KD, distributed on PyPI as \texttt{sail-kd} and imported as \texttt{kd}.
It requires Python 3.11 or later.
The source code is available at \href{https://github.com/scientific-artificial-intelligence-lab/KD}{\nolinkurl{github.com/scientific-artificial-intelligence-lab/KD}}.
The following command installs LegoSR together with the components that LegoSR Agent needs:

\par\nopagebreak\noindent\begin{minipage}{\linewidth}
\begin{lstlisting}[language=bash]
python -m pip install "sail-kd[agent]"
\end{lstlisting}
\end{minipage}\par

PySR needs the additional \texttt{pysr} component, which downloads a Julia runtime on first use; \texttt{sail-kd[agent,pysr]} installs both.
PyTorch is installed as a dependency.
On a machine without a GPU, or with a driver that does not support the newest CUDA build, the matching PyTorch build should be installed first, for example the CPU build:

\par\nopagebreak\noindent\begin{minipage}{\linewidth}
\begin{lstlisting}[language=bash]
python -m pip install torch --index-url https://download.pytorch.org/whl/cpu
\end{lstlisting}
\end{minipage}\par

Two commands check the installation:

\par\nopagebreak\noindent\begin{minipage}{\linewidth}
\begin{lstlisting}[language=bash]
python -c "import kd; print(kd.__version__)"
kd-agent --help
\end{lstlisting}
\end{minipage}\par

\subsection{Configuring the language model}
\label[subappendix]{app:llm}

LegoSR Agent and the LLM4ED method call a language model through an API.
Searches with the other methods do not need this step.
The command \texttt{kd-agent setup} asks for four values.
It saves them in \texttt{\textasciitilde/.config/\allowbreak kd/\allowbreak agent.json}, a file that only the user can read:

\par\nopagebreak\noindent\begin{minipage}{\linewidth}
\begin{lstlisting}[language=bash]
$ kd-agent setup
Endpoint (openai, anthropic, local, nvidia, bigmodel): openai
Base URL (leave empty for a preset default): https://api.deepseek.com
Model (leave empty for a preset default): deepseek-flash
API key:
Configuration saved to ~/.config/kd/agent.json
\end{lstlisting}
\end{minipage}\par

The endpoint \texttt{openai} accepts any service with an OpenAI-compatible API, here the DeepSeek model used in \cref{sec:experiments}.
The endpoint \texttt{anthropic} is for Anthropic-compatible APIs, and \texttt{local} for a model that runs on the same machine through Ollama.
The key is typed at a hidden prompt and is never passed as a command argument, so it does not appear in the shell history.
Environment variables override the saved values, which is convenient on a computing cluster:

\par\nopagebreak\noindent\begin{minipage}{\linewidth}
\begin{lstlisting}[language=bash]
export KD_AGENT_ENDPOINT=openai
export KD_AGENT_BASE_URL=https://api.deepseek.com
export KD_AGENT_MODEL=deepseek-flash
export KD_AGENT_API_KEY=<your key>
\end{lstlisting}
\end{minipage}\par

LLM4ED keeps its own settings.
In Python, its base URL and model are set in \texttt{kd.Llm4edConfig}, and its key is read from the environment variable \texttt{OPENAI\_API\_KEY}.
LegoSR Agent uses LLM4ED only after \texttt{kd-agent setup -{}-llm4ed} has saved a separate endpoint and key for it; the key of the agent itself is never passed to LLM4ED.

\subsection{Discovery with Python code}
\label[subappendix]{app:python}

The Python interface has three steps: load the data, run a discovery method and inspect the result.
The example below loads the Burgers data bundled with LegoSR and runs SGA-PDE for 30 generations with 20 candidate equations each.
In this and the following examples, a comment line after a \texttt{print} call shows its output:

\par\nopagebreak\noindent\begin{minipage}{\linewidth}
\begin{lstlisting}[language=Python]
import kd

data = kd.load("burgers")
model = kd.Model(algorithm="sga", generations=30,
                 population=20, seed=0)
model.fit(data)
print(model.result_.equation)
# u_t = -u*u_x + 0.1002*u_xx
model.report("out/burgers/")
\end{lstlisting}
\end{minipage}\par

\begin{figure}[!htb]
\centering
\includegraphics[width=\textwidth]{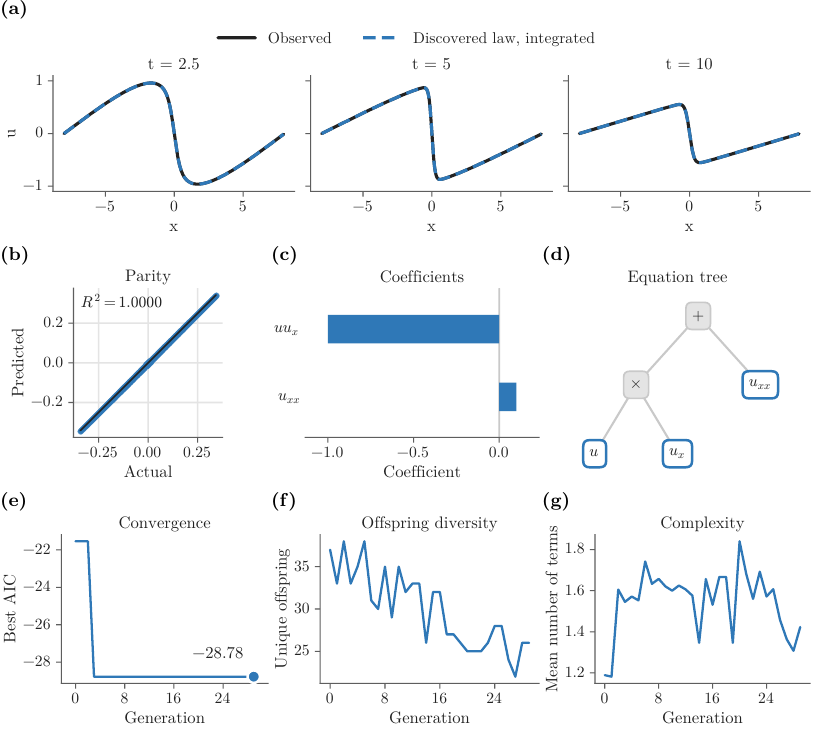}
\caption{Some of the figures drawn by LegoSR for this example. (a) Observed field and the integrated equation at $t = 2.5$, 5 and 10. (b) $u_t$ from the data against the right-hand side of the discovered equation. (c, d) Coefficients and expression tree of the discovered equation. (e) Lowest Akaike information criterion (AIC), the score of SGA-PDE, found up to each generation. (f, g) Plots added by SGA-PDE: number of distinct expressions and mean number of terms among the offspring of each generation.}
\label{fig:report-gallery}
\end{figure}

\texttt{model.result\_} holds the discovered equation and the search history, and \texttt{model.report} writes figures and an HTML report to the given folder.
The figures include the observed field and the forward error shown in \cref{fig:quickstart}, a coefficient chart, residual and parity plots, the expression tree of the equation and the search history (\cref{fig:report-gallery}).
Some methods add their own plots, such as the per-generation statistics of SGA-PDE.
Each figure is also available as a function in \texttt{kd.viz.plots}, such as \texttt{plot\_parity(model.result\_)}, for use in one's own figures.

The online documentation shows every figure of a full report (\href{https://scientific-artificial-intelligence-lab.github.io/kd/viz/gallery/}{\nolinkurl{scientific-artificial-intelligence-lab.github.io/kd/viz/gallery}}), and the notebook \href{https://github.com/scientific-artificial-intelligence-lab/KD/blob/trunk/examples/notebooks/getting_started.ipynb}{\texttt{getting\_started.ipynb}} runs a shorter version of this example step by step.

Another discovery method is selected through the \texttt{algorithm} argument, while the data and the analysis code stay the same.
Settings such as \texttt{population} belong to SGA-PDE and are left out.
\Cref{tab:solver-options} lists the name of each method in LegoSR, the data it accepts, the source of its derivatives and what it needs beyond the base installation.
For example, PySINDy with its default settings recovers the same equation:

\par\nopagebreak\noindent\begin{minipage}{\linewidth}
\begin{lstlisting}[language=Python]
model = kd.Model(algorithm="pysindy")
model.fit(data)
print(model.result_.equation)
# u_t = -u*u_x + 0.1002*u_xx
\end{lstlisting}
\end{minipage}\par

\begin{table}[!htb]
\centering
\caption{Equation discovery methods in LegoSR. Grids hold fields on regular grids; tabular data hold input variables and a target, as in algebraic SR and ODE problems. Derivatives come from finite differences of the data or from a neural surrogate fitted to them.}
\label{tab:solver-options}
\footnotesize
\begin{tabularx}{\textwidth}{@{}l l >{\raggedright}p{2.5cm} >{\raggedright}p{3.1cm} >{\raggedright\arraybackslash}X@{}}
\toprule
Method & \texttt{algorithm} & Data & Derivatives & Additional requirement \\
\midrule
PySINDy & \texttt{pysindy} & Grids,\newline tabular data & Finite differences & None \\
DLGA-PDE & \texttt{dlga} & Grids & Neural surrogate & None \\
SGA-PDE & \texttt{sga} & Grids,\newline tabular data & Finite differences or neural surrogate & None \\
PySR & \texttt{pysr} & Grids,\newline tabular data & Finite differences & \texttt{pysr} component; Julia downloaded on first use \\
DISCOVER & \texttt{discover} & Grids,\newline tabular data & Finite differences or neural surrogate & None \\
LLM4ED & \texttt{llm4ed} & Grids,\newline tabular data & Finite differences & Language-model API key (\cref{app:llm}) \\
EqGPT & \texttt{eqgpt} & Grids,\newline scattered points & Finite differences or neural surrogate & Pretrained weights, downloaded on first use \\
\bottomrule
\end{tabularx}
\end{table}

A candidate equation can also be checked without running a discovery method.
\texttt{kd.refit\_terms} fits the coefficients of given terms to the data and reports the normalized residual $J$ of \cref{eq:residual} (attribute \texttt{nmse}), and \texttt{kd.forward\_error} integrates the equation from the first observed time and returns its relative $L_2$ error:

\par\nopagebreak\noindent\begin{minipage}{\linewidth}
\begin{lstlisting}[language=Python]
fit = kd.refit_terms(data, ["u*u_x", "u_xx"])
print([f"{c:.4g}" for c in fit.coefficients], f"{fit.nmse:.1e}")
# ['-1', '0.1002'] 1.1e-05
err = kd.forward_error(data, ["u*u_x", "u_xx"], fit.coefficients)
print(f"{err.value:.1e}")
# 1.8e-04
\end{lstlisting}
\end{minipage}\par

Data from a file are loaded with the same function.
For an NPZ file that stores the coordinates \texttt{x} and \texttt{t} and the field \texttt{u} as arrays, the mappings name the arrays and \texttt{periodic} marks a periodic axis:

\par\nopagebreak\noindent\begin{minipage}{\linewidth}
\begin{lstlisting}[language=Python]
mine = kd.load("my_data.npz",
               coords={"x": "x", "t": "t"},
               fields={"u": "u"},
               periodic=["x"])
# ODE trajectories: map only the time, coords={"t": "t"},
#   and list every state variable in fields.
# Algebraic data: lhs names the target, e.g. lhs="y";
#   the other columns become the input variables.
\end{lstlisting}
\end{minipage}\par

\texttt{kd.inspect\_file} lists the arrays in a file when their names are not known.
Other formats, such as CSV, HDF5 and XLSX, are loaded with the same function, and the notebook \href{https://github.com/scientific-artificial-intelligence-lab/KD/blob/trunk/examples/notebooks/your_own_data.ipynb}{\texttt{your\_own\_data.ipynb}} builds a dataset directly from NumPy arrays.

A Sketch adds known physics to the search (\cref{sec:framework-representation}).
It is useful when part of the equation is known, for example a diffusion term or a conservation form, or when earlier searches agree on some terms.
A Sketch has three parts.
Anchored terms must appear in the equation, and their coefficients are fitted.
Pinned terms are fixed together with their coefficients.
Holes are left for the discovery method to fill, and each hole can limit the number of terms, the derivative order, the operators and the variables it may use.
The Sketch below anchors the diffusion term $u_{xx}$ and leaves a hole for one further term:

\par\nopagebreak\noindent\begin{minipage}{\linewidth}
\begin{lstlisting}[language=Python]
sketch = kd.Sketch(anchored=["u_xx"],
                   holes=[kd.Hole(max_terms=1)])
model = kd.Model(algorithm="sga", seed=0)
model.fit(data, sketch=sketch)
print(model.result_.equation)
# u_t = -u*u_x + 0.1002*u_xx
\end{lstlisting}
\end{minipage}\par

SGA-PDE searches for the term in the hole and fits the coefficient of the anchored term, and LegoSR checks that the final equation contains the anchored term.
The Sketch of \cref{eq:sketch} instead pins the advection term with its coefficient and leaves a hole for at most two terms with derivatives up to second order:

\par\nopagebreak\noindent\begin{minipage}{\linewidth}
\begin{lstlisting}[language=Python]
sketch = kd.Sketch(pinned={"u*u_x": -1.0},
                   holes=[kd.Hole(max_terms=2, max_deriv_order=2)])
model = kd.Model(algorithm="sga", seed=0)
model.fit(data, sketch=sketch)
print(model.result_.equation)
# u_t = -u*u_x + 0.1002*u_xx
\end{lstlisting}
\end{minipage}\par

LegoSR moves the pinned term to the left-hand side, so SGA-PDE fits $u_t + u\,u_x$ with the terms in the hole; the pinned term is restored in the final equation.
Any of the seven methods accepts the same Sketch.

\subsection{Checking candidate equations on noisy data}
\label[subappendix]{app:checks}

With noisy data, finite-difference derivatives are often dominated by noise, so a candidate equation should also be checked in ways that need no derivatives of the data.
The example below uses the Allen--Cahn data bundled with LegoSR, which follow $u_t = 0.003\,u_{xx} + u - u^3$.
Like the Chafee--Infante equation of \cref{sec:prior_guided}, this equation combines diffusion with a cubic reaction term.
\texttt{kd.add\_noise} adds Gaussian noise whose standard deviation is 5\% of that of the field.
SGA-PDE with finite-difference derivatives then returns an equation without any derivative:

\par\nopagebreak\noindent\begin{minipage}{\linewidth}
\begin{lstlisting}[language=Python]
clean = kd.load("allen-cahn")
data = kd.add_noise(clean, 0.05, seed=0)
model = kd.Model(algorithm="sga", generations=30,
                 population=20, seed=0)
model.fit(data)
print(model.result_.equation)
# u_t = -0.1216*x**3
\end{lstlisting}
\end{minipage}\par

Suppose that earlier searches or physical reasoning suggest the terms $u_{xx}$, $u$ and $u^3$.
Refitted with finite differences, these terms leave almost all of the variation of $u_t$ unexplained ($J = 0.998$).
Weak-form regression integrates the equation against smooth test functions and needs no derivatives of the data~\cite{messenger2021wsindy}.
With \texttt{method="weak"}, the same terms leave a weak-form residual of 0.0084, with coefficients within 0.5\% of the true values:

\par\nopagebreak\noindent\begin{minipage}{\linewidth}
\begin{lstlisting}[language=Python]
terms = ["u_xx", "u", "u**3"]
fd = kd.refit_terms(data, terms)
weak = kd.refit_terms(data, terms, method="weak")
print(f"{fd.nmse:.3g}  {weak.nmse:.2g}")
# 0.998  0.0084
print([f"{c:.4g}" for c in weak.coefficients])
# ['0.002994', '0.9997', '-0.9955']
\end{lstlisting}
\end{minipage}\par

\texttt{kd.ablate\_terms} tests whether each term is needed, as in \cref{fig:chafee_checks}b.
It drops one term at a time, refits the others and reports how much the residual rises relative to that of the full equation.
The test below adds an advection term $u_x$ that does not belong to the equation:

\par\nopagebreak\noindent\begin{minipage}{\linewidth}
\begin{lstlisting}[language=Python]
check = kd.ablate_terms(data, terms + ["u_x"], method="weak")
for row in check.rows:
    print(f"{row.dropped_term:5s} {row.relative_rise:.3g}")
# u_xx  67
# u     103
# u**3  81.8
# u_x   9.34e-06
\end{lstlisting}
\end{minipage}\par

Dropping $u_x$ leaves the residual almost unchanged, whereas dropping any other term raises it by 67--103 times its value for the full equation, so $u_x$ is removed.
Finally, the equation is integrated forward and compared with the field.
Because the noise was added here, the noise-free field is available, and the recovered equation reproduces it with a relative $L_2$ error of 0.27\%:

\par\nopagebreak\noindent\begin{minipage}{\linewidth}
\begin{lstlisting}[language=Python]
err = kd.forward_error(clean, terms, weak.coefficients)
print(f"{err.value:.2%}")
# 0.27%
\end{lstlisting}
\end{minipage}\par

Several discovery methods can also take derivatives from a neural network fitted to the noisy field (\cref{tab:solver-options}).
The notebook \href{https://github.com/scientific-artificial-intelligence-lab/KD/blob/trunk/examples/notebooks/noisy_burgers_walkthrough.ipynb}{\texttt{noisy\_burgers\_\allowbreak walkthrough.ipynb}} compares finite-difference and network derivatives on Burgers data with 15\% noise.

\subsection{Discovery with LegoSR Agent}
\label[subappendix]{app:agent}

The command \texttt{kd-agent run} hands the whole task to LegoSR Agent.
In \cref{sec:quickstart}, the agent receives the first 161 time frames of the bundled Burgers data, and the remaining frames are kept for checking.
These frames are written to a file with NumPy:

\par\nopagebreak\noindent\begin{minipage}{\linewidth}
\begin{lstlisting}[language=Python]
import numpy as np
import kd

train = kd.load("burgers").window("t", stop=8.0)
np.savez("burgers.npz", x=train.get_coords("x").numpy(),
         t=train.get_coords("t").numpy(),
         u=train.get_field("u").numpy())
\end{lstlisting}
\end{minipage}\par

A file that holds a time array \texttt{t}, a spatial coordinate \texttt{x} and one field on their grid loads without further options, so the command of \cref{sec:quickstart} runs the agent on it:

\par\nopagebreak\noindent\begin{minipage}{\linewidth}
\begin{lstlisting}[language=bash]
kd-agent run --data burgers.npz "Find the equation governing this data."
\end{lstlisting}
\end{minipage}\par

With \texttt{-{}-dataset burgers} in place of \texttt{-{}-data burgers.npz}, the agent receives all frames of the bundled data.
For other files, \texttt{-{}-load-options} takes the arguments of \texttt{kd.load} (\cref{app:python}) as a JSON object:

\par\nopagebreak\noindent\begin{minipage}{\linewidth}
\begin{lstlisting}[language=bash]
kd-agent run --data my_data.npz \
  --load-options '{"coords": {"x": "x", "t": "t"},
                   "fields": {"u": "u"}, "periodic": ["x"]}' \
  "Find the equation governing this data."
\end{lstlisting}
\end{minipage}\par

Each run has its own folder under \texttt{./kd-agent-runs/}, and the option \texttt{-{}-workspace} reopens a folder in a later session.
The first section of the file \texttt{report.md} in this folder gives the submitted equation and its coefficients.
For one run of LegoSR Agent on the Burgers data above, it reads:

\par\nopagebreak\noindent\begin{minipage}{\linewidth}
\begin{lstlisting}
## Submitted law

LHS: `u_t`

$$
u_{t} = - 1.0000744328048 u u_{x} + 0.100207836692516 u_{xx}
$$

| Term | Coefficient (full precision) |
|---|---|
| `diff2_x(u)` | 0.10020783669251626 |
| `mul(u_x, u)` | -1.0000744328048001 |
\end{lstlisting}
\end{minipage}\par

The report then lists every search with its method, settings and status, including searches that failed, and gives the reason for the final choice.
In this run, the agent searched with PySINDy and then with SGA-PDE, and it chose the SGA-PDE equation for ``the lowest NMSE and cleanest coefficients, independently confirming the pysindy fit.''
\texttt{kd-agent chat} takes the same data options and opens an interactive session, in which the researcher can discuss the search with the agent and steer it.

A researcher gives a Sketch to the agent as a file.
After the Sketch is saved with \texttt{sketch.save("sketch.json")}, the agent runs every search under it and fills the holes:

\par\nopagebreak\noindent\begin{minipage}{\linewidth}
\begin{lstlisting}[language=bash]
kd-agent run --data my_data.npz --sketch sketch.json \
  "Find the equation governing this data."
\end{lstlisting}
\end{minipage}\par

In \texttt{kd-agent chat}, the researcher can also describe the known terms in words, and the agent writes the Sketch.

Besides its built-in method cards and workflow guidance, LegoSR Agent can read usage notes for the methods and the Experience Library from files:

\par\nopagebreak\noindent\begin{minipage}{\linewidth}
\begin{lstlisting}[language=bash]
kd-agent run --data my_data.npz --advice advice.json \
  --memory experience.json "Find the equation governing this data."
\end{lstlisting}
\end{minipage}\par

The Experience Library is built from searches graded against known equations, for example on benchmark problems with reference solutions:

\par\nopagebreak\noindent\begin{minipage}{\linewidth}
\begin{lstlisting}[language=bash]
kd-agent experience build --graded-runs graded.jsonl \
  --out experience.json
\end{lstlisting}
\end{minipage}\par

The documentation of LegoSR Agent in the repository describes the format of the graded searches (\href{https://github.com/scientific-artificial-intelligence-lab/KD/tree/trunk/pdeagent}{\nolinkurl{pdeagent}}).
The \href{https://scientific-artificial-intelligence-lab.github.io/kd/guide/controller/}{online guide} describes the other options of \texttt{kd-agent}, and an \href{https://scientific-artificial-intelligence-lab.github.io/kd/guide/agent-example/}{example session} saves a search and resumes it.

\section{Graphical user interface}
\label[appendix]{app:webui}

\begin{figure}[!htbp]
\centering
\includegraphics[width=\textwidth]{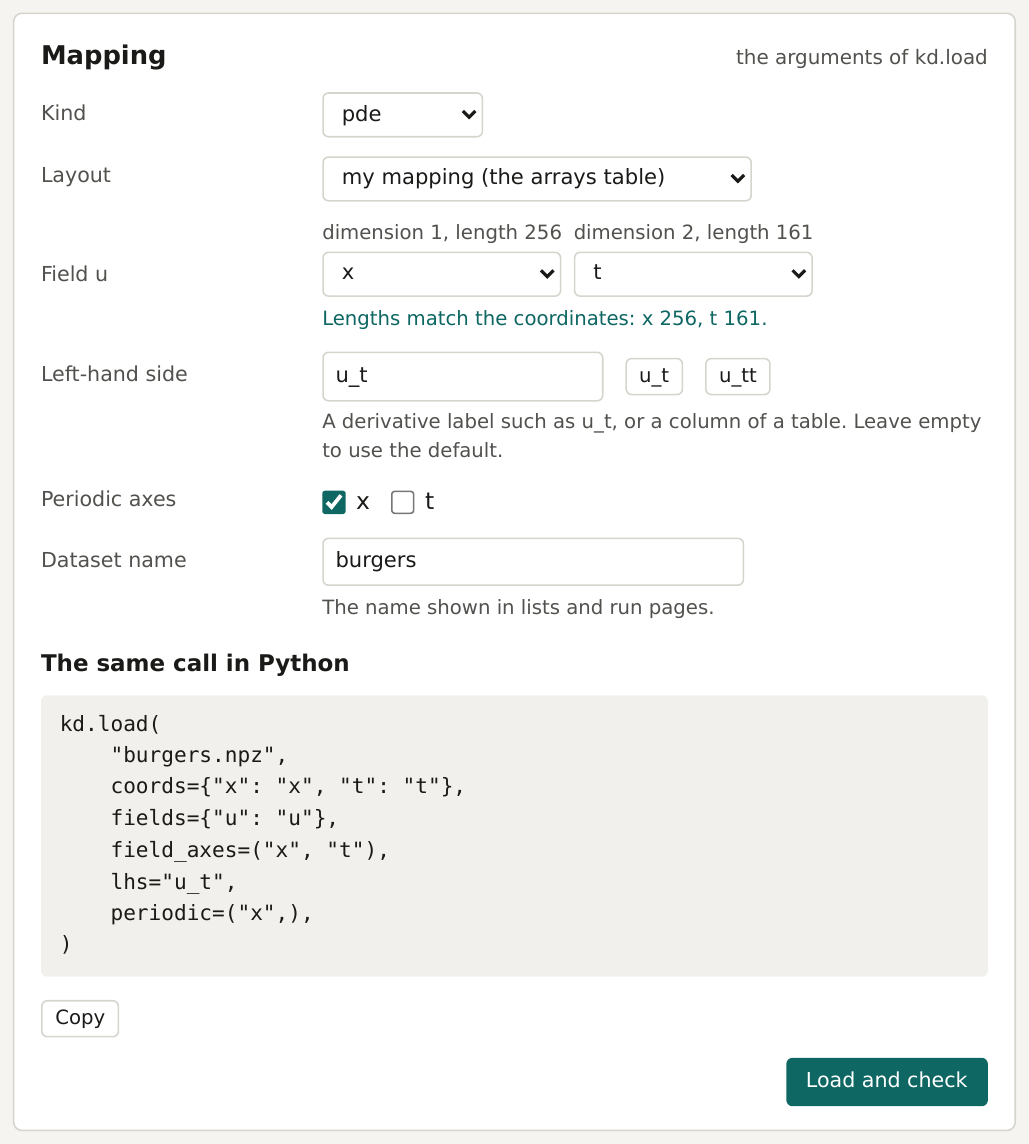}
\caption{Loading the Burgers data of \cref{app:python} from an NPZ file in the graphical interface. The form maps the arrays to the coordinates and the field, and the page shows the equivalent \texttt{kd.load} call with every argument written out, including those left at their defaults in \cref{app:python}.}
\label{fig:webui-data}
\end{figure}

The steps of \cref{app:python} are also available in a graphical user interface that runs in a local browser.
It is installed with the \texttt{ui} component, \texttt{sail-kd[ui]}, and the command \texttt{kd-ui} starts it and prints the address to open.
A researcher loads a file and maps its arrays to coordinates and fields, and the page shows the equivalent \texttt{kd.load} call (\cref{fig:webui-data}).
The page then shows which discovery methods accept the data and gives the reason for each refusal.

\begin{figure}[!htbp]
\centering
\includegraphics[width=\textwidth]{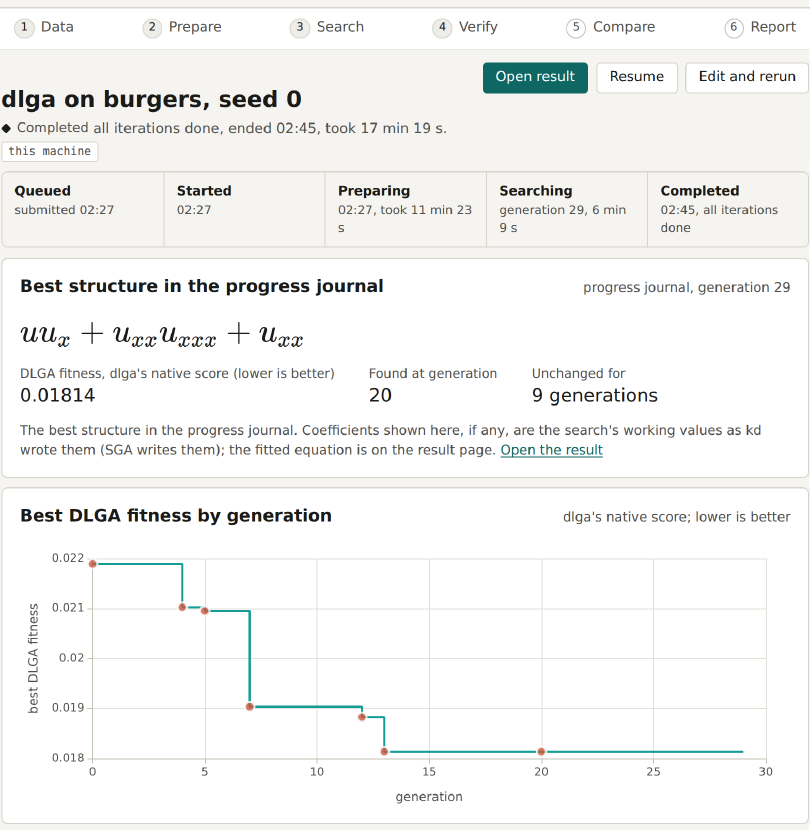}
\caption{Monitoring a DLGA-PDE search on the Burgers data (seed 0, 30 generations). Selected panels from the completed run page show navigation, run controls, the stage timeline, the best candidate structure and the native score history. Dots mark recorded best-candidate updates. The progress journal displays term structures; fitted coefficients are reported on the result page.}
\label{fig:webui-run}
\end{figure}

\begin{figure}[!htbp]
\centering
\includegraphics[width=\textwidth]{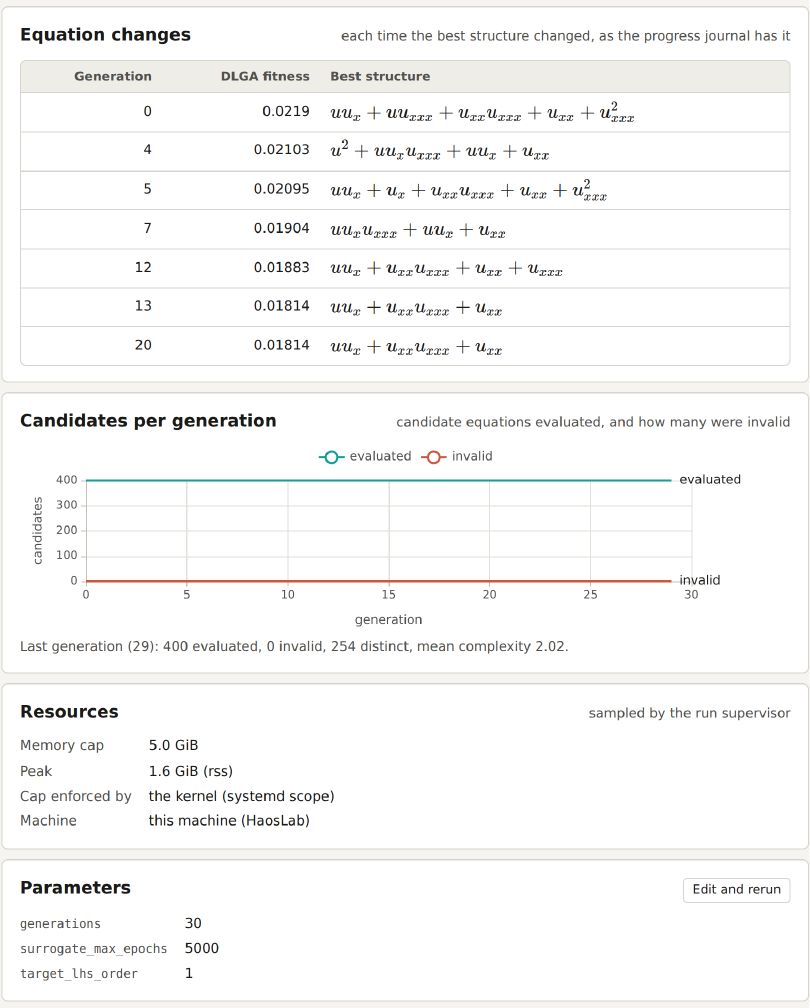}
\caption{Further panels from the run page in \cref{fig:webui-run}: recorded best-candidate updates, evaluated and invalid candidates at each generation, resource usage and search parameters.}
\label{fig:webui-run-details}
\end{figure}

The settings form of each discovery method is generated from the parameters that the method declares.
While a search runs, its page shows the best candidate structure so far, the best native score at each generation and the best-candidate updates recorded in the progress journal (\cref{fig:webui-run,fig:webui-run-details}).
The same page reports candidate counts, resource usage and search parameters.

\begin{figure}[!htbp]
\centering
\includegraphics[width=\textwidth]{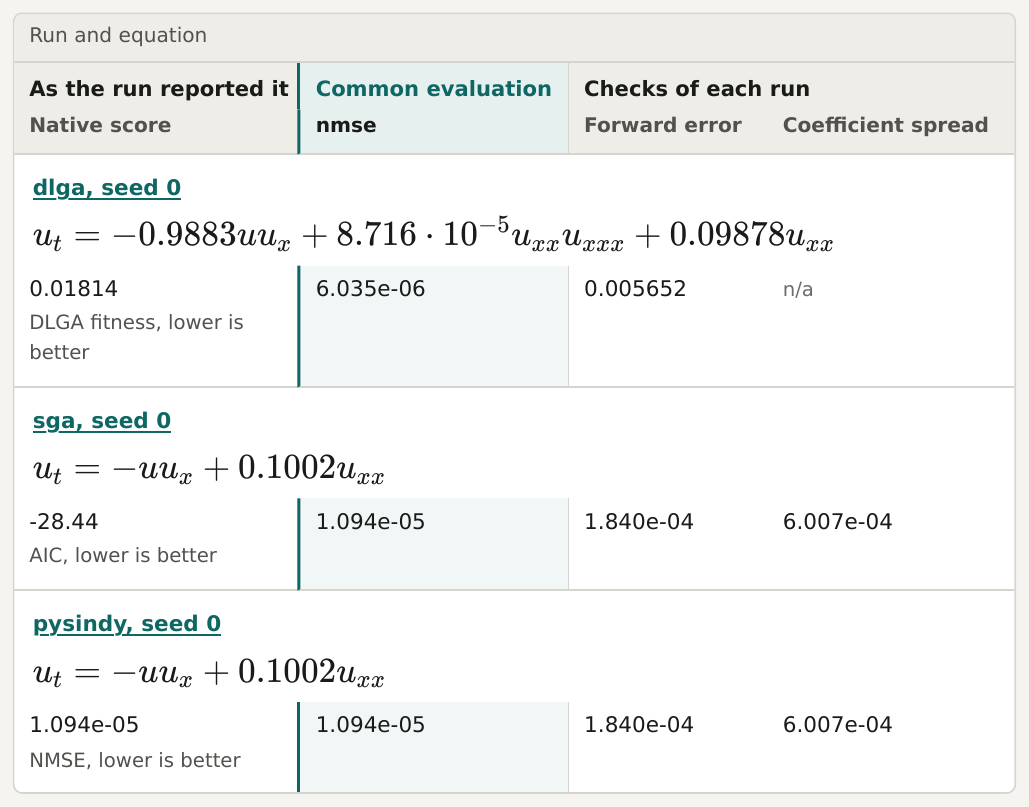}
\caption{DLGA-PDE, SGA-PDE and PySINDy runs on the Burgers data compared in the graphical interface: each method's native score, the residual $J$ (nmse) from the common evaluation, the forward error and the coefficient spread from the subdomain test. Native scores follow each method's own definition, while LegoSR recomputes $J$ in the same way for every run. Runs are labelled by the \texttt{algorithm} names of \cref{tab:solver-options}.}
\label{fig:webui-compare}
\end{figure}

The page of each result reports three separate checks: the forward error, the drop-one-term test and the subdomain test of \cref{sec:framework-evaluation}.
Runs of several methods on the same data are compared on one page, which lists for each run the native score, the residual $J$ from the common evaluation of \cref{sec:framework-evaluation}, the forward error and the coefficient spread from the subdomain test (\cref{fig:webui-compare}).
Run folders written by Python scripts or by LegoSR Agent can be imported and viewed in the same pages.

\clearpage
\section{Benchmark problems}
\label[appendix]{app:benchmarks}

The algebraic SR problems are the 100 Feynman equations and 20 bonus equations of the AI Feynman database~\cite{udrescu2020aifeynman} and the Strogatz systems of SRBench~\cite{lacava2021srbench}, as described in \cref{sec:exp-setup}.
\Cref{tab:ode-problems} lists the 19 ODE systems, which give 40 target equations, and \cref{tab:pde-problems} lists the 25 PDE problems.
The last seven ODE systems are also the Strogatz systems of the algebraic SR benchmark; here their derivatives must be estimated from trajectories.
PDE domains are given as space $\times$ time, and a half-open interval marks a periodic axis.
Grid sizes refer to the complete data before the training--test split.
In \cref{tab:pde-problems}, the citation after each problem name gives the source of its data.
The KdV data keep every second spatial point of the $512 \times 201$ source grid.
The Chafee--Infante data use the reaction term $-u + u^3$ of their source, whereas the classical Chafee--Infante equation $u_t = u_{xx} + \lambda (u - u^3)$~\cite{chafee1974bifurcation} has the opposite sign.
The linear advection data are generated with the recipe of MDBench~\cite{ziaeibideh2026mdbench}, which follows PDEBench~\cite{takamoto2022pdebench}.

\begin{table}[!htb]
\centering
\caption{ODE systems.}
\label{tab:ode-problems}
\footnotesize
\renewcommand{\arraystretch}{1.2}
\begin{tabularx}{\textwidth}{@{}>{\raggedright}p{4.4cm} >{\raggedright\arraybackslash}X@{}}
\toprule
System & Equations \\
\midrule
Harmonic oscillator & $\dot{x} = v$, \quad $\dot{v} = -1.69\,x$ \\
Damped harmonic oscillator & $\dot{x} = v$, \quad $\dot{v} = -1.69\,x - 0.26\,v$ \\
Damped pendulum & $\dot{\theta} = v$, \quad $\dot{v} = -1.44\sin\theta - 0.08\,v$ \\
Duffing oscillator & $\dot{x} = v$, \quad $\dot{v} = -x - 0.8\,x^3 - 0.12\,v$ \\
Van der Pol oscillator & $\dot{x} = v$, \quad $\dot{v} = 2\,(1 - x^2)\,v - x$ \\
Lorenz system & $\dot{x} = 10\,(y - x)$, \quad $\dot{y} = x\,(28 - z) - y$, \quad $\dot{z} = x y - \tfrac{8}{3} z$ \\
Lotka--Volterra predator--prey & $\dot{x} = 1.1\,x - 0.4\,x y$, \quad $\dot{y} = 0.1\,x y - 0.4\,y$ \\
SIR epidemic model & $\dot{S} = -0.9\,S I$, \quad $\dot{I} = 0.9\,S I - 0.25\,I$, \quad $\dot{R} = 0.25\,I$ \\
FitzHugh--Nagumo & $\dot{v} = v - v^3/3 - w + 0.5$, \quad $\dot{w} = 0.08\,(v + 0.7 - 0.8\,w)$ \\
Brusselator & $\dot{x} = 1 - 4\,x + x^2 y$, \quad $\dot{y} = 3\,x - x^2 y$ \\
Michaelis--Menten kinetics & $\dot{s} = -1.5\,s/(0.7 + s)$ \\
Robertson kinetics & $\dot{x} = -0.04\,x + 10^4 y z$, \quad $\dot{y} = 0.04\,x - 10^4 y z - 3\times 10^7 y^2$, \quad $\dot{z} = 3\times 10^7 y^2$ \\
\midrule
Bacterial respiration & $\dot{x} = 20 - x - x y/(1 + 0.5\,x^2)$, \quad $\dot{y} = 10 - x y/(1 + 0.5\,x^2)$ \\
Bar magnets & $\dot{x} = 0.5\sin(x - y) - \sin x$, \quad $\dot{y} = 0.5\sin(y - x) - \sin y$ \\
Glider & $\dot{x} = -0.05\,x^2 - \sin y$, \quad $\dot{y} = x - \cos y / x$ \\
Lotka--Volterra competition & $\dot{x} = 3\,x - 2\,x y - x^2$, \quad $\dot{y} = 2\,y - x y - y^2$ \\
Predator--prey (Holling type II) & $\dot{x} = x\,\bigl(4 - x - y/(1 + x)\bigr)$, \quad $\dot{y} = y\,\bigl(x/(1 + x) - 0.075\,y\bigr)$ \\
Shear flow & $\dot{x} = \cot y \cos x$, \quad $\dot{y} = (\cos^2 y + 0.1\sin^2 y)\sin x$ \\
Van der Pol (Li\'enard form) & $\dot{x} = 10\,\bigl(y - (x^3 - x)/3\bigr)$, \quad $\dot{y} = -0.1\,x$ \\
\bottomrule
\end{tabularx}
\end{table}

\clearpage
\begingroup
\footnotesize
\renewcommand{\arraystretch}{1.2}
\setlength{\tabcolsep}{4pt}
\begin{longtable}{@{}>{\raggedright\arraybackslash}p{3.3cm} >{\raggedright\arraybackslash}p{\dimexpr\textwidth-9.1cm-6\tabcolsep\relax} >{\raggedright\arraybackslash}p{3.35cm} >{\raggedright\arraybackslash}p{2.45cm}@{}}
\caption{PDE problems.}\label{tab:pde-problems}\\
\toprule
Problem & Equation & Domain & Grid \\
\midrule
\endfirsthead
\caption[]{PDE problems (continued).}\\
\toprule
Problem & Equation & Domain & Grid \\
\midrule
\endhead
\midrule
\multicolumn{4}{r}{\textit{Continued on the next page}}\\
\endfoot
\bottomrule
\endlastfoot
Burgers (sine initial state)~\cite{chen2022sgapde} & $u_t = -u\,u_x + 0.1\,u_{xx}$ & $[-8, 8) \times [0, 10]$ & $256 \times 201$ \\
Burgers (Gaussian initial state)~\cite{rudy2017pdefind} & $u_t = -u\,u_x + 0.1\,u_{xx}$ & $[-8, 8) \times [0, 10]$ & $256 \times 101$ \\
Burgers (shock)~\cite{raissi2019pinns} & $u_t = -u\,u_x + (0.01/\uppi)\,u_{xx}$ & $[-1, 1] \times [0, 0.99]$ & $256 \times 100$ \\
2D Burgers~\cite{xu2025eqgpt} & $u_t = -u\,u_x - u\,u_y + 0.01\,u_{xx} + 0.01\,u_{yy}$ & $[-1, 1]^2 \times [0, 1.98]$ & $101 \times 51 \times 100$ \\
KdV~\cite{chen2022sgapde} & $u_t = -u\,u_x - 0.0025\,u_{xxx}$ & $[-1, 1) \times [0, 1]$ & $256 \times 201$ \\
Kuramoto--Sivashinsky~\cite{raissi2018deephpm} & $u_t = -u\,u_x - u_{xx} - u_{xxxx}$ & $[-10, 10) \times [0, 50]$ & $511 \times 251$ \\
Allen--Cahn~\cite{xu2025eqgpt} & $u_t = 0.003\,u_{xx} + u - u^3$ & $[-1, 1) \times [0, 10]$ & $256 \times 201$ \\
Chafee--Infante~\cite{chen2022sgapde} & $u_t = u_{xx} - u + u^3$ & $[0, 3] \times [0.1, 0.498]$ & $301 \times 200$ \\
Fisher--KPP~\cite{du2024llm4ed} & $u_t = 0.02\,u_{xx} + 10\,u - 10\,u^2$ & $[-1, 1] \times [0, 1]$ & $201 \times 101$ \\
Fisher--KPP with nonlinear diffusion~\cite{du2024llm4ed} & $u_t = 0.02\,u\,u_{xx} + 0.02\,u_x^2 + 10\,u - 10\,u^2$ & $[-1, 1] \times [0, 1]$ & $201 \times 101$ \\
Heat~\cite{du2024llm4ed} & $u_t = 0.05\,u_{xx}$ & $[0, 10) \times [0, 10]$ & $200 \times 201$ \\
Nonlinear diffusion~\cite{chen2022sgapde} & $u_t = u\,u_{xx} + u_x^2$ & $[1, 2] \times [0, 0.5]$ & $100 \times 251$ \\
Linear advection~\cite{ziaeibideh2026mdbench} & $u_t = -0.1\,u_x$ & $[0, 1) \times [0, 2]$ & $256 \times 201$ \\
Convection--diffusion (I)~\cite{xu2025eqgpt} & $u_t = -u_x + 0.25\,u_{xx}$ & $[0, 2] \times [0, 1]$ & $256 \times 100$ \\
Convection--diffusion (II)~\cite{xu2021dlpde} & $u_t = -u_x + 0.25\,u_{xx}$ & $[0, 30] \times [0.1, 15]$ & $120 \times 150$ \\
Advection--diffusion with velocity $1/x$~\cite{chen2022sgapde} & $u_t = -u_x/x + 0.25\,u_{xx}$ & $[1, 2] \times [0, 1]$ & $100 \times 251$ \\
2D advection--diffusion~\cite{both2021deepmod} & $u_t = 0.25\,u_x + 0.5\,u_y + 0.5\,u_{xx} + 0.5\,u_{yy}$ & $[-5, 5]^2 \times [3, 9]$ & $51 \times 51 \times 61$ \\
Klein--Gordon~\cite{xu2025eqgpt} & $u_{tt} = 0.5\,u_{xx} - 5\,u$ & $[-1, 1] \times [0, 3]$ & $201 \times 201$ \\
Wave~\cite{xu2025eqgpt} & $u_{tt} = u_{xx}$ & $[0, \uppi] \times [0, 2\uppi]$ & $161 \times 321$ \\
Euler--Bernoulli beam~\cite{stephany2021pderead} & $u_{tt} = -0.1\,u_{xxxx}$ & $[-5, 5] \times [0, 20]$ & $401 \times 401$ \\
Nonlinear Schr\"odinger ($u$)~\cite{raissi2018deephpm} & $u_t = -0.5\,v_{xx} - u^2v - v^3$ & $[-5, 5) \times [0, \uppi/2]$ & $512 \times 501$ \\
Nonlinear Schr\"odinger ($v$)~\cite{raissi2018deephpm} & $v_t = 0.5\,u_{xx} + u^3 + uv^2$ & $[-5, 5) \times [0, \uppi/2]$ & $512 \times 501$ \\
2D reaction--diffusion ($u$), generated~\cite{rudy2017pdefind} & $u_t = 0.1(u_{xx} + u_{yy}) + u - u^3 - uv^2 + u^2v + v^3$ & $[-5.9375, 5.9375]^2$ $\times [0, 10]$ & $77 \times 77 \times 101$ \\
2D reaction--diffusion ($v$), generated~\cite{rudy2017pdefind} & $v_t = 0.1(v_{xx} + v_{yy}) + v - u^3 - uv^2 - u^2v - v^3$ & $[-5.9375, 5.9375]^2$ $\times [0, 10]$ & $77 \times 77 \times 101$ \\
Variable-coefficient diffusion (generated) & $u_t = \partial_x(K\,u_x)$; $K(x) = 1 + 0.5\sin(2\uppi x)$ & $[0, 1] \times [0, 0.04]$ & $101 \times 161$ \\
\end{longtable}
\endgroup

\clearpage

\bibliographystyle{elsarticle-num}
\bibliography{refs}

\end{document}